\documentclass[10pt,journal,compsoc]{IEEEtran}

\usepackage[nocompress]{cite}

\usepackage{url}
\usepackage{ragged2e}
\usepackage{epsfig}
\usepackage{graphicx}
\usepackage{amsmath} % define this before the line numbering.
\usepackage{amssymb}
\usepackage{subfigure}
\usepackage{algorithm}
\usepackage{algorithmicx}
\usepackage{algpseudocode}
\usepackage{graphics}
\usepackage{mathrsfs}
\usepackage{threeparttable}
\usepackage{color}
\usepackage[normalem]{ulem}
\usepackage{multirow}
\usepackage{float}
\usepackage{amsfonts}
\usepackage{bm}
\usepackage{array}
\usepackage[table]{xcolor}
\usepackage{colortbl}
\usepackage{pifont}
\usepackage{diagbox}
\usepackage{rotating}
\usepackage{booktabs}
\usepackage{overpic}
\usepackage{textcomp}
\usepackage{contour}
\usepackage{enumitem}
\usepackage{stfloats}
\usepackage{threeparttable}
\usepackage{makecell}
\usepackage[colorlinks=true,breaklinks=true,urlcolor=magenta]{hyperref}

\definecolor{myRed}{RGB}{219, 68, 55}
\definecolor{myGreen}{RGB}{15, 157, 88}
\definecolor{myBlue}{RGB}{66, 133, 244}

\newcommand{\jy}[1]{{\textcolor{black}{#1}}}

\def\eg{{e.g.}}
\def\ie{{i.e.}}

\def\etal{{et al.}}

\graphicspath{{./Imgs/}}
\DeclareGraphicsExtensions{.jpg,.pdf,.png}

\begin{document}

\title{Effectiveness Assessment of Recent Large Vision-Language Models}

\author{
Yao Jiang$^\dag$,~
Xinyu Yan$^\dag$,~
Ge-Peng Ji,~
Keren Fu*,~
Meijun Sun,~
Huan Xiong*,~\\
Deng-Ping Fan$^\ddag$,~
Fahad Shahbaz Khan \\
\IEEEcompsocitemizethanks{
% \IEEEcompsocthanksitem Yao Jiang, Xinyu Yan, Huan Xiong, Deng-Ping Fan, and Fahad Shahbaz Khan are with Mohamed bin Zayed University of Artificial Intelligence, Abu Dhabi, UAE.
\IEEEcompsocthanksitem Yao Jiang is with Mohamed Bin Zayed University of Artificial Intelligence, Abu Dhabi, UAE, and Sichuan University, Chengdu, China.
\IEEEcompsocthanksitem Xinyu Yan is with Mohamed Bin Zayed University of Artificial Intelligence, Abu Dhabi, UAE, and Tianjin University, Tianjin, China.
\IEEEcompsocthanksitem Ge-Peng Ji is with Australian National University, Canberra, Australia.
\IEEEcompsocthanksitem Keren Fu is with Sichuan University, Chengdu, China.
\IEEEcompsocthanksitem Meijun Sun is with Tianjin University, Tianjin, China.
\IEEEcompsocthanksitem Huan Xiong is with Mohamed Bin Zayed University of Artificial Intelligence, Abu Dhabi, UAE, and Harbin Institute of Technology, Harbin, China.
\IEEEcompsocthanksitem Deng-Ping Fan is with Nankai University, Tianjin, China.
\IEEEcompsocthanksitem Fahad Shahbaz Khan is with Mohamed Bin Zayed University of Artificial Intelligence, Abu Dhabi, UAE.
% \IEEEcompsocthanksitem Huan Xiong and Fahad Shahbaz Khan are with Mohamed Bin Zayed University of Artificial Intelligence, Abu Dhabi, UAE.
% \IEEEcompsocthanksitem Qi Ma is with Nankai University, Tianjin, China.
% \IEEEcompsocthanksitem Ge-Peng Ji is with Australian National University, Canberra, Australia.
\IEEEcompsocthanksitem $\dag$ Two authors contribute equally. $\ddag$ Deng-Ping Fan is the project lead. * Corresponding authors: Keren Fu (fkrsuper@scu.edu.cn) and Huan Xiong (huan.xiong.math@gmail.com). Work was done while Yao Jiang and Xinyu Yan were MBZUAI visiting scholars.
}
}

% \markboth{IEEE TRANSACTIONS ON PATTERN ANALYSIS AND MACHINE INTELLIGENCE}%
% {Chou \MakeLowercase{\textit{et al.}}: Colonoscopy review}

\IEEEtitleabstractindextext{%
\begin{abstract} \justifying
The advent of large vision-language models (LVLMs) represents a remarkable advance in the quest for artificial general intelligence. However, the model's effectiveness in both specialized and general tasks warrants further investigation. This paper endeavors to evaluate the competency of popular LVLMs in specialized and general tasks, respectively, aiming to offer a comprehensive understanding of these novel models. To gauge their effectiveness in specialized tasks, we employ six challenging tasks in three different application scenarios: natural, healthcare, and industrial. These six tasks include salient/camouflaged/transparent object detection, as well as polyp detection, skin lesion detection, and industrial anomaly detection. 
We examine the performance of three recent open-source LVLMs, including MiniGPT-v2, LLaVA-1.5, and Shikra, on both visual recognition and localization in these tasks. 
Moreover, we conduct empirical investigations utilizing the aforementioned LVLMs together with GPT-4V, assessing their multi-modal understanding capabilities in general tasks including object counting, absurd question answering, affordance reasoning, attribute recognition, and spatial relation reasoning. 
Our investigations reveal that these LVLMs demonstrate limited proficiency not only in specialized tasks but also in general tasks. We delve deep into this inadequacy and uncover several potential factors, including limited cognition in specialized tasks, object hallucination, text-to-image interference, and decreased robustness in complex problems. We hope that this study can provide useful insights for the future development of LVLMs, helping researchers improve LVLMs for both general and specialized applications.
\end{abstract}

}

\maketitle

\IEEEdisplaynontitleabstractindextext

\IEEEpeerreviewmaketitle

\section{Introduction}\label{sec:introduction}
The emergence of large language models (LLMs)~\cite{openai2023gpt3,touvron2023llama} has sparked a revolution in the field of natural language processing, owing to their promising generalization and reasoning capabilities. Motivated by this progress, researchers have pioneered the development of powerful large vision-language models (LVLMs)~\cite{liu2023visualllava,chen2023minigptv2,openai2023gpt4v}, leveraging the impressive capabilities of LLMs to enhance comprehension of visual semantics. This advance particularly improves model performance in complex vision-language tasks~\cite{chen2023minigptv2,fu2023challengerGemini,qin2023goodbard}, and represents a major step toward artificial general intelligence (AGI). 
\jy{AGI refers to intelligent systems that are capable of solving any task that can be performed by humans or animals. Generally, tasks performed by humans can be divided into general and specialized tasks according to whether special domain knowledge is required.
Therefore, the capabilities of LVLMs can be categorized into these two aspects accordingly, and both of them are essential for LVLMs on the path toward AGI.
}

Recently, numerous studies have assessed and investigated the general and specialized capabilities of LVLMs~\cite{qin2023goodbard,fu2023challengerGemini,zhang2023exploringAD,tang2023generalizationcod,10261199,gu2023anomalygpt,fu2023mme}. Qin~\etal~\cite{qin2023goodbard} conducted empirical studies encompassing various general tasks, such as object detection and counting to evaluate the visual understanding capabilities of Google Bard. Fu~\etal~\cite{fu2023mme} introduced a comprehensive evaluation benchmark to assess the perceptual and cognitive capabilities of recent LVLMs on general tasks (\eg, optical character recognition and object counting). Zhang~\etal~\cite{zhang2023exploringAD} explored the potential of GPT-4V~\cite{openai2023gpt4v} in visual anomaly detection, while Tang~\etal~\cite{tang2023generalizationcod} generalized Shikra~\cite{chen2023shikra} to challenging camouflaged object detection scenarios without training.
However, as these studies primarily focus on evaluating the general capabilities of LVLMs~\cite{qin2023goodbard,fu2023challengerGemini,fu2023mme} or exploring the effectiveness of a particular LVLM in a specialized domain~\cite{zhang2023exploringAD,tang2023generalizationcod,10261199,gu2023anomalygpt}, there is a lack of quantitative analysis regarding the performance of recent LVLMs in a diverse range of specialized tasks, leading to an insufficient understanding of their capabilities.

\begin{figure*}[t]
\footnotesize
  \centering
  \includegraphics[width=\linewidth]{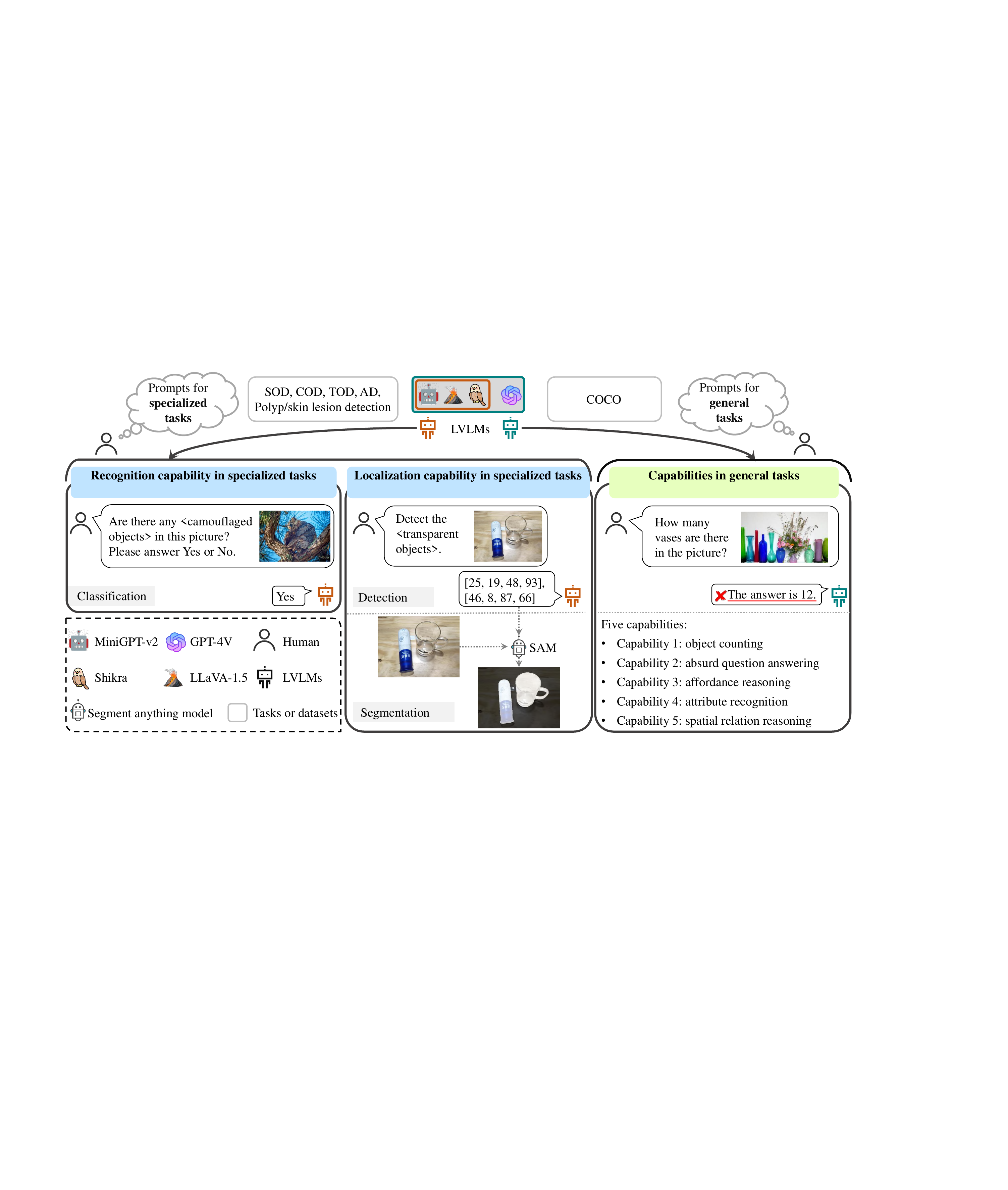}
   \put(-96, 158){\footnotesize(\S~\ref{sec:quali_other})}
  \put(-434, 158){\footnotesize(\S~\ref{sec:perception})}
  \put(-260, 158){\footnotesize(\S~\ref{sec:localization})}
      \caption{Overall diagram of our evaluation platform. We evaluate the recent LVLMs in both specialized and general tasks using tailored prompts, with and without specifying object types. The specialized tasks include salient object detection (SOD), transparent object detection (TOD), camouflaged object detection (COD), polyp detection, skin lesion detection, as well as industrial anomaly detection (AD). The evaluation is realized by conducting recognition (\S~\ref{sec:perception}) and localization (\S~\ref{sec:localization}) under these tasks, and three recent open-source LVLMs (MiniGPT-v2~\cite{chen2023minigptv2}, LLaVA-1.5~\cite{liu2023improvedllava1point5}, and Shikra~\cite{chen2023shikra}) are tested.
      Besides, empirical investigations are conducted on the COCO~\cite{lin2014microsoftCOCO} dataset to reflect the capabilities of LVLMs in general tasks (\S~\ref{sec:quali_other}), including object counting, absurd question answering, affordance reasoning, attribute recognition, and spatial relation reasoning.
      Examples are presented in each figure group, where ``$<$...$>$'' indicates a placeholder that can be replaced with other words/phrases in different tasks.}
 % \vspace{-10pt}
  \label{fig:teaser_image}
\end{figure*}

In this paper, we conduct a comprehensive assessment of several recent open-source LVLMs, spanning a diverse array of challenging specialized and general tasks. 
Our evaluation platform is illustrated in Fig.~\ref{fig:teaser_image}. 
To evaluate the ability of LVLMs to perform specialized tasks, we select three recent open-source LVLMs (MiniGPT-v2~\cite{chen2023minigptv2}, LLaVA-1.5~\cite{liu2023improvedllava1point5}, and Shikra~\cite{chen2023shikra}) and conduct quantitative assessment on six challenging specialized tasks in three different application scenarios: natural, healthcare, and industrial.
\jy{For natural scenarios, we select salient object detection (SOD)~\cite{song20233d,fu2021siamese,fu2020jl}, transparent object detection (TOD)~\cite{xie2020segmentingTrans10K}, and camouflaged object detection (COD)~\cite{fan2022concealed,ji2022fast}, as these tasks involve targets that are increasingly rare in real-life and possess progressively complex characteristics, thereby presenting distinct challenges to LVLMs. In the field of healthcare, the effectiveness of LVLMs is evaluated by skin lesion detection~\cite{codella2018skinISIC2017} and polyp detection~\cite{tajbakhsh2015automatedColonDB}, which show prominent and slightly weaker visual features, respectively. Besides, anomaly detection (AD)~\cite{bergmann2021mvtec}, a vital task in industrial scenarios, is also selected for assessment. In academia, these six tasks come with tailored datasets and cover broad specialized domains, thereby enabling comprehensive evaluation of specialized capabilities of LVLMs.
}
As illustrated in Fig.~\ref{fig:teaser_image}, given inherent challenges posed by these tasks in terms of recognizing and localizing target objects, we employ tailored prompts to assess the recognition (\S~\ref{sec:perception}) and localization (\S~\ref{sec:localization}) capabilities of the models.
Furthermore, we conduct empirical investigations on a universal dataset (COCO~\cite{lin2014microsoftCOCO}) that is free from domain-specific expertise. We abstain from specifying particular object types (``camouflaged'', ``transparent'', or else) in prompts, aiming to explore multi-modal understanding capabilities (\S~\ref{sec:quali_other}) of the above-mentioned models and GPT-4V in general tasks (\ie, object counting, absurd question answering, affordance reasoning, attribute recognition, and spatial relation reasoning).
\jy{
 The assessed prominent LVLMs, include MiniGPT-v2~\cite{chen2023minigptv2}, LLaVA-1.5~\cite{liu2023improvedllava1point5}, Shikra~\cite{chen2023shikra}, and GPT-4V~\cite{openai2023gpt4v}, all of which have garnered significant research attention as key players in the field. Among them, three accessible open-source models, \ie, MiniGPT-v2, LLaVA-1.5, and Shikra, are selected to ensure feasibility and reproducibility of the evaluation in specialized tasks.}

Our investigations reveal that while these models show strong potential for specialized tasks, they still exhibit suboptimal performance and limited cognitive capabilities. This reveals their inadequate \jy{transfer} ability in this particular context.
Performance issues are further magnified by typical weaknesses of LVLMs such as object hallucination, text-to-image interference, and decreased robustness in complex problems. In addition to the shortcomings revealed in specialized tasks, these models also show significant room for improvement in general tasks, particularly in object counting, spatial reasoning, and absurd question answering. 

In summary, the main contributions of this paper are three-fold: 
(1) We construct an evaluation platform comprising six specialized tasks and five general tasks to assess the effectiveness of LVLMs. 
(2) On the evaluation platform, we evaluate the specialized capabilities of three recent open-source LVLMs and also the general capabilities of four LVLMs. 
(3) We analyze their performance and limitations for both specialized and general tasks, and discuss the future development and application of LVLMs.

\section{Recognition via LVLMs in Specialized Tasks}\label{sec:perception}
When LVLMs are applied in these specialized tasks, recognizing these target objects is a crucial step, which reflects models' global understanding of such tasks and directly influences their effectiveness. Therefore, we first conduct quantitative evaluation of their recognition capabilities on the aforementioned six specialized tasks. Subsequently, we carry out additional tests to delve into failure cases and gain further insights.

\subsection{Quantitative Investigation}\label{sec:quanti_cls}
% \vspace{-5pt}
\subsubsection{Experimental Setup} 
Recognition in specialized tasks involves determining the existence of targets and classifying them. 
The first evaluation of recognition capabilities is to judge object existence, requiring models to answer either ``Yes'' or ``No'' to questions like ``Are there any $<$camouflaged objects$>$ in the picture? Please answer Yes or No.'' as demonstrated in Fig.~\ref{fig:teaser_image}. The placeholder ``$<$...$>$'' in the queries denotes flexible words/phrases that can be substituted in different tasks, such as ``polyps'' in polyp detection. 
The evaluation considers two different setups: the full set, which includes both positive and negative samples, and the positive set, which includes only positive samples.

Beyond the first evaluation, we delve deeper into the fine-grained recognition ability of LVLMs by asking them to categorize targets. Our method is to prompt LVLMs to designate the most suitable category for a target object from a pre-defined set of potential categories  (w/ vocabulary).
Within this experiment, the questions like ``Which of the following is the most likely category for the camouflaged object in the picture? `seahorse, mantis, spider ...' '' are used. The pre-defined set contains all categories that appear in the dataset. 
Besides, another evaluation is considered, featuring an open-vocabulary inquiry without giving a pre-defined set (w/o vocabulary). In this test, a straightforward question like ``What is the camouflaged object in the picture?'' is used.

\jy{The versions of LLavA-1.5~\cite{liu2023improvedllava1point5}, Shikra~\cite{chen2023shikra}, and MiniG-PT-v2~\cite{chen2023minigptv2} that are equipped with language models of $\sim$7 billion parameters are selected for evaluation. All configurations of each model are set as default during evaluation. Since all tests in this paper are based on the above configurations, we will not mention again in the following sections.
}

\subsubsection{Metrics}
As for the first evaluation, accuracy ($\mathcal{A}$) is employed to measure the performance of LVLMs in judging object existence, while the probability of positive responses (responses indicating ``yes'') on the full set is also reported for reference.
$\mathcal{A}$ and the probability of positive responses ($\mathcal{Y}$) can be formulated as follows:
\begin{align}
    \mathcal{A} = \frac{\rm TP+TN}{\rm TP+FP+TN+FN}, \\
    \mathcal{Y} = \frac{\rm TP+FP}{\rm TP+FP+TN+FN},
\end{align}
where ${\rm TP, FP, TN, FN}$ denote true positive, false positive, true negative, and false negative, respectively.  

For fine-grained recognition, LVLMs typically select categories from a pre-defined set when available, enabling direct matching with labels for assessing correctness. However, in the absence of such a set, the generated categories exhibit significant variation, posing challenges in directly evaluating correctness through class matching.
Hence, we utilize accuracy ($\mathcal{A}^*$) and semantic similarity ($\mathcal{S}$)~\cite{conti2023vocabularyfree} to measure the performance in these two settings, respectively. The former quantifies the fraction of responses that contain correct category names, while the latter quantifies the semantic similarity between responses and ground truth labels. Considering that LVLMs may occasionally generate similar categories not included in the pre-defined set, $\mathcal{S}$ is also employed to evaluate the performance of the w/ vocabulary setting.

\subsubsection{Benchmark Datasets}\label{sec:cls_dataset}
A total of 10 datasets from SOD (DUTS~\cite{wang2017learningDUTS} and SOC~\cite{fan2018salientSOC}), COD (COD10K~\cite{fan2022concealed}), TOD (Trans10K~\cite{xie2020segmentingTrans10K}), polyp detection (ColonDB~\cite{tajbakhsh2015automatedColonDB}, ETIS~\cite{silva2014towardETIS}, and CP-CHILD-B~\cite{wang2020improvedchildb}), skin lesion detection (ISIC~\cite{codella2018skinISIC2017}), and AD (MVTec AD~\cite{bergmann2021mvtec} and VisA~\cite{zou2022spotVisA}) are employed to evaluate the performance of LVLMs in determining
the existence of targets. 
Among these datasets, SOC, COD10K, CP-CHILD-B, MVTec AD, and VisA, which contain both positive and negative samples, are used to construct the full set, while the remaining datasets are utilized to form the positive set.
The proportions of positive samples in SOC, COD10K, CP-CHILD-B, MVTec AD, and VisA are 50\%, 50.7\%, 25\%, 72.9\%, and 55.5\%, respectively.

COD10K, the only dataset that provides category labels for each target, is utilized to evaluate the fine-grained recognition ability of LVLMs. Since judging target existence in negative samples is certainly challenging for LVLMs, we exclude the interference and use only the positive samples of COD10K to more accurately evaluate the fine-grained recognition ability.

\subsubsection{Result Analyses and Discussions}
Evaluation results of existence determination on the full set and positive set, and fine-grained recognition are detailed in Tables~\ref{tab:lvlm_cls_positive_and_negative}-\ref{tab:fgcod_cls}. The absence of negative samples leads to $\rm TN = 0$ and $\rm FP = 0$, and hence $\mathcal{A}$ in Table~\ref{tab:lvlm_cls_all_positive} is equivalent to $\mathcal{Y}$ in Table~\ref{tab:lvlm_cls_positive_and_negative}. Three observations from these results are as follows.

\textbf{Over-positive issue}.
From the results in Table~\ref{tab:lvlm_cls_positive_and_negative} and the proportion of positive samples in each dataset (in \S~\ref{sec:cls_dataset}), we can observe that these models consistently yield a greater proportion of positive responses ($\mathcal{Y}$) compared to the proportion of positive samples. Especially on SOC and CP-CHILD-B, these LVLMs generally achieve $\mathcal{Y}$ higher than 0.9, while the proportions of positive samples in these datasets are only 50\% and 25\%. 
This indicates that the models tend to give positive responses, which is further proved on the positive sets in Table~\ref{tab:lvlm_cls_all_positive}, where extremely high scores on $\mathcal{A}$ (\eg, 1.000) are achieved (particularly for LLaVA-1.5). The reason behind this phenomenon could be that most of the samples learned by these LVLMS during the training are positive image-text pairs, which makes them over-positive and thus have a tendency to answer ``yes'' to the questions~\cite{li2023evaluatingobjecthallucination,xu2023lvlm}.

\begin{table*}[t!]
\footnotesize
% \tiny
\centering
\caption{Experimental results for three LVLMs regarding the presence of targets on the full sets. We present the probability of positive answers ($\mathcal{Y}$, representing the percentage of ``yes''). The highest accuracy ($\mathcal{A}$) score is highlighted in bold.}
\label{tab:lvlm_cls_positive_and_negative}
\tabcolsep 18pt
\begin{tabular*}{\textwidth}{lccccc}
            \toprule \hline
           \multirow{3}{*}[-1.0ex]{Model} & \multicolumn{2}{c}{Natural scenes} & \multicolumn{1}{c}{Healthcare}            & \multicolumn{2}{c}{Industrial scenes} \\ \cmidrule(lr{0pt}){2-3} \cmidrule(lr{0pt}){4-4} \cmidrule(lr{0pt}){5-6} 
           & SOC\cite{fan2018salientSOC}       &COD10K\cite{fan2022concealed}      & CP-CHILD-B\cite{wang2020improvedchildb}     & MVTec AD\cite{bergmann2021mvtec}          & VisA\cite{zou2022spotVisA}           \\ \cmidrule(lr{0pt}){2-6}
  & $\mathcal{A}$/$\mathcal{Y}$    & $\mathcal{A}$/$\mathcal{Y}$ & $\mathcal{A}$/$\mathcal{Y}$ & $\mathcal{A}$/$\mathcal{Y}$ & $\mathcal{A}$/$\mathcal{Y}$ \\ \midrule
MiniGPT-v2\cite{chen2023minigptv2} & 0.513/0.987        & 0.580/0.909        &   0.250/0.990      & 0.695/0.874         & 0.543/0.875        \\
 LLaVA-1.5~\cite{liu2023improvedllava1point5}&\textbf{0.618}/0.883&\textbf{0.776}/0.427&   0.268/0.983      & \textbf{0.750}/0.979& 0.562/0.993        \\
Shikra~\cite{chen2023shikra}     & 0.528/0.973        & 0.535/0.053        &\textbf{0.285}/0.945& 0.728/0.562         &\textbf{0.617}/0.251\\ \hline \bottomrule
\end{tabular*}
\end{table*}

\begin{table*}[t!]
\footnotesize
% \tiny
\centering
\caption{Experimental results for three LVLMs regarding the presence of targets on the positive sets. The highest accuracy score is marked in bold. Given the absence of negative samples in the positive set, resulting in ${\rm TN=0}$ and ${\rm FP=0}$, the metric $\mathcal{A}$ in this table is equivalent to $\mathcal{Y}$.}
\label{tab:lvlm_cls_all_positive}
\tabcolsep 22pt
\begin{tabular*}{\textwidth}{lccccc}
            \toprule \hline
           \multirow{3}{*}[-1.0ex]{Model} & \multicolumn{2}{c}{Natural scenes} & \multicolumn{3}{c}{Healthcare}            \\ \cmidrule(lr{0pt}){2-3} \cmidrule(lr{0pt}){4-6} 
           & DUTS\cite{wang2017learningDUTS}       & Trans10K\cite{xie2020segmentingTrans10K}   & ColonDB~\cite{tajbakhsh2015automatedColonDB}      & ETIS\cite{silva2014towardETIS}         & ISIC~\cite{codella2018skinISIC2017}          \\ \cmidrule(lr{0pt}){2-6}
             & $\mathcal{A}$    & $\mathcal{A}$ & $\mathcal{A}$ & $\mathcal{A}$ & $\mathcal{A}$ \\ \midrule
MiniGPT-v2~\cite{chen2023minigptv2} & 0.853         & 0.964        & 0.824        & 0.847        & 0.952        \\
LLaVA-1.5~\cite{liu2023improvedllava1point5} & 0.999         &\textbf{1.000}&\textbf{1.000}&\textbf{0.985}&\textbf{1.000}  \\
Shikra~\cite{chen2023shikra}     & \textbf{1.000}& 0.988        & 0.968        & 0.954        & 0.998        \\ \hline \bottomrule
\end{tabular*}
\end{table*}

\begin{table}[t]
\footnotesize
\centering
\caption{Quantitative results of three LVLMs for classifying camouflaged objects. The best results are marked in bold. We conduct classification solely on the positive samples within \jy{COD10K~\cite{fan2022concealed}}.}
\label{tab:fgcod_cls}
\tabcolsep 3pt
\begin{tabular*}{0.48\textwidth}{lcccc}
\toprule \hline
             Setting   &      Metric        & MiniGPT-v2\cite{chen2023minigptv2} & Shikra\cite{chen2023shikra} & LLaVA-1.5\cite{liu2023improvedllava1point5}          \\ \midrule
\textit{w/ vocabulary} & $\mathcal{A}^*$ & 0.285    & 0.154  & \textbf{0.436}    \\
\textit{w/ vocabulary} & $\mathcal{S}$  & 0.567    & 0.545  & \textbf{0.673}    \\
\textit{w/o vocabulary} & $\mathcal{S}$ & 0.607    & 0.608  & \textbf{0.655}    \\ \hline \bottomrule
\end{tabular*}
\end{table}

\textbf{Limited performance in determining existence}.
Though notably high accuracy ($\mathcal{A}$) in Table~\ref{tab:lvlm_cls_all_positive} are achieved by LVLMs, the inclusion of negative samples results in an overall decrease in accuracy. As shown in Table~\ref{tab:lvlm_cls_positive_and_negative}, most accuracies drop below 0.7, indicating an inadequate recognition ability of LVLMs in determining the existence of targets, particularly in the case where negative samples are presented.  
Among these models, LLaVA-1.5 shows better recognition capabilities for camouflaged objects, achieving higher accuracy ($\mathcal{A}$) while obtaining $\mathcal{Y}$ scores that are close to the proportions of positive samples in COD10K. In contrast,
Shikra shows extremely bad results (on $\mathcal{Y}$) on COD10K due to its frequent misclassification of positive samples, indicating its less capability in recognizing camouflaged objects.

% \textbf{Struggling in camouflaged object perception}.
\textbf{Struggling with classifying camouflaged objects}.
The results in Table~\ref{tab:fgcod_cls} clearly demonstrate that these LVLMs struggle with classifying camouflaged objects. Although LLaVA-1.5 achieves the highest scores, its performance is still unsatisfactory.  
The unsatisfactory performance could be attributed to various factors. 
First, these models may face challenges in identifying camouflaged objects that closely resemble the background, as indicated by their unsatisfactory recognition accuracy in Table~\ref{tab:lvlm_cls_positive_and_negative}.
Second, the category of camouflaged objects may lie beyond the models' domain of knowledge, hindering their capability to match objects with their categories accurately.
Additionally, the extended length of the prompt, stemming from the incorporation of the pre-defined set, may impede the model's comprehension. 
This aligns with the results in Table~\ref{tab:fgcod_cls}, where MiniGPT-v2 and Shikra demonstrate improved performance ($\mathcal{S}$) when the pre-defined set is excluded (i.e. w/o vocabulary), as opposed to when the vocabulary is provided (i.e. w/ vocabulary).

\subsection{Uncovering Insights into Failure Cases}\label{sec:quali_cls}
\begin{figure*}[t!]
\footnotesize
% \vspace{-20pt}
\includegraphics[width=\linewidth]{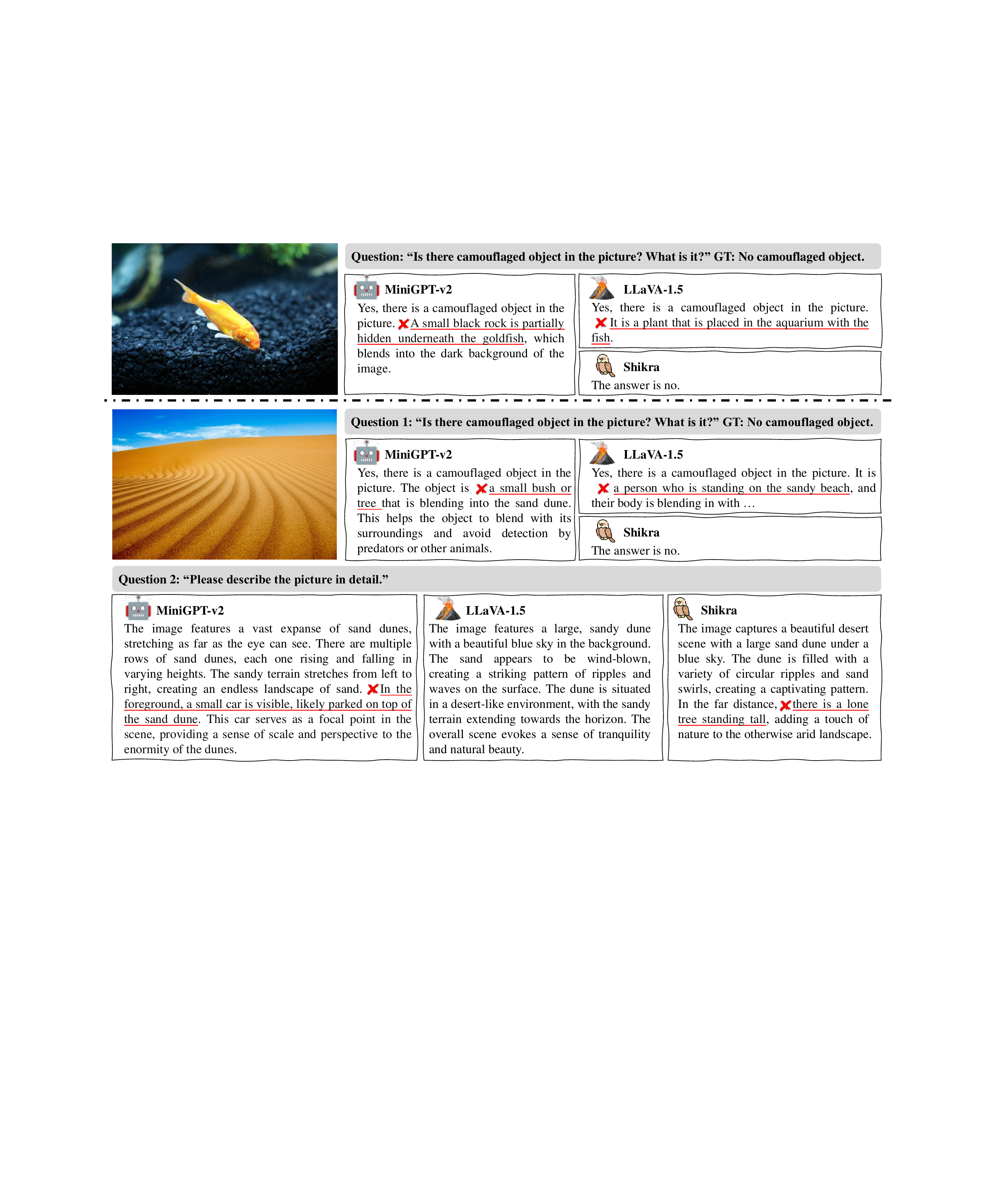}
\caption{Responses of three LVLMs regarding the perception of camouflaged objects on negative samples. Incorrect responses are underlined in red and marked with crosses.}
\label{fig:qualitative_cls}
\end{figure*}

Recalling that these models encounter challenges in differentiating negative samples, so we conduct tests on representative negative samples to gain insight into the potential causes of this phenomenon. 
LVLMs are prompted to provide additional descriptions or reasoning when determining the existence of targets.
The results are illustrated in Fig.~\ref{fig:qualitative_cls}, deriving three potential factors.

\textbf{Limited cognition towards specific attributes}.
As illustrated in the first example of Fig.~\ref{fig:qualitative_cls}, when presented with the question ``Is there camouflaged object in the picture? What is it?'', MiniGPT-v2 erroneously recognizes the ``small black rock'' as a camouflaged object, while LLaVA-1.5 misclassifies a ``plant'' as such.
These models classify rocks and plants as camouflaged objects just because of their visual resemblance to the surroundings, indicating their limited cognition about camouflage. This phenomenon also occurs in other specialized tasks, \eg, anomaly detection, implying their limited cognition on special object types.

\textbf{Object hallucinations}.
Object hallucination, which involves imagining objects in the response but not present in the image~\cite{li2023evaluatingobjecthallucination,cui2023holistic}, could impact the recognition capability of LVLMs in specialized tasks.
For instance, as demonstrated by the answers to ``Is there a camouflaged object in the picture? What is it?'' in the second example of Fig.~\ref{fig:qualitative_cls}, LLaVA-1.5 states that ``a person is standing on the sandy beach'', while MiniGPT-v2 mentions the presence of ``small bush or tree''. 
These objects may affect the recognition of targets~\cite{tang2023generalizationcod}, resulting in decreased recognition performance in determining object presence.

\textbf{Text-to-image interference}. 
The inadequate performance in determining the presence of targets may also be attributed to text-to-image interference, which originates from the textual prompts supplied to the models~\cite {cui2023holistic}. As shown in the second example in Fig.~\ref{fig:qualitative_cls}, when prompted with ``Please describe the picture in detail'', LLaVA-1.5 provides an accurate description of the image. However, when prompted with ``Is there a camouflaged object in the picture? What is it?'', the mention of the ``camouflaged object'' in the prompt may interfere with the answers, resulting in hallucination and misjudgment from LLaVA-1.5.

\subsection{Summary}
Sect. 2 evaluates the recognition performance of MiniGPT-v2~\cite{chen2023minigptv2}, LLaVA-1.5~\cite{liu2023improvedllava1point5}, and Shikra~\cite{chen2023shikra} in various specialized tasks. 
\jy{Among them, LLaVA-1.5 generally shows better recognition ability in both existence determination and object classification.
}
However, quantitative analyses indicate that while these models exhibit certain cognitive capabilities in various specialized tasks without domain-specific fine-tuning, their recognition performance requires further enhancement. When directly applied to these tasks, they still achieve limited cognition and understanding of specialized domains.
Apart from such limited cognition, other typical weaknesses of LVLMs, as revealed in qualitative investigations, such as object hallucination and text-to-image interference, are likely to result in inferior performance.

\section{Localization via LVLMs in Specialized Tasks}\label{sec:localization}
In this section, we assess the localization capabilities of three LVLMs on the six specialized tasks, and further explore their strengths and limitations through additional qualitative tests.

\subsection{Quantitative Investigation}\label{sec:quanti_det}

\subsubsection{Experimental Setup}
Recent LVLMs have demonstrated a remarkable visual grounding capability as they can locate objects with bounding boxes (bboxes) that are specified in language prompts. This capability makes it feasible to apply these models to the specialized tasks described above. 
To achieve this goal, we employ a two-step methodology consisting of detection followed by segmentation.
Specifically, as illustrated in Fig.~\ref{fig:teaser_image}, we initially prompt LVLMs to provide bounding boxes for a particular type of objects (\eg, transparent objects) with a question like ``Detect the $<$transparent objects$>$.'', yielding  detection of targets. Subsequently, the predicted bounding boxes are used as further prompts to the segment anything model (SAM)~\cite{kirillov2023segmentanything} to perform fine segmentation.
\jy{Given the potential presence of multiple boxes in a picture, we first employ SAM to generate a separate mask for each box and then merge these results using the Boolean OR operation to obtain the final segmentation result.
} 
The SAM with the ViT-H backbone~\cite{dosovitskiy2020imageVIT} is employed as the default in all the experiments.
We also conduct segmentation using ground truth bounding boxes, which serve as the upper bound of segmentation performance.

\subsubsection{Metrics}
As mentioned previously, we perform detection followed by segmentation to utilize these models for specialized tasks. Therefore, during evaluation, we assess their localization capabilities by evaluating their performance in both detection and segmentation. To evaluate the detection results, three widely used detection metrics (\ie, $\rm Precision$, $\rm Recall$, and $\rm F1$ with an intersection-over-union (IoU) threshold of 0.5~\cite{electronics10030279}) are adopted. Additionally, three segmentation metrics (mean absolute error ($M$)~\cite{Perazzi2012SaliencyFC}, S-measure ($S_\alpha$)~\cite{Fan2017StructureMeasureAN}, and maximum F-measure ($F_\beta$)~\cite{Achanta2009FrequencytunedSR}) are employed to assess segmentation performance.
It should be noted that since these models solely predict bounding boxes without providing corresponding confidence values, we exclude those common metrics like average precision ($\rm AP$)~\cite{electronics10030279} in anomaly detection.

\begin{table*}[t!]
\footnotesize
\centering
\caption{Detection and segmentation results of MiniGPT-v2, LLaVA-1.5, and Shikra in natural scenarios. The symbols $\uparrow$/$\downarrow$ indicate that a higher/lower score is better, while the highest scores are marked in bold. 
The \textcolor{gray}{upper bound} (on ground truth bounding boxes) of detection and segmentation via LVLMs in diverse specialized tasks is marked in gray.}
\label{tab:lvlm_dec_and_seg_natural}
% \tabcolsep 10pt
% \begin{tabular*}{\textwidth}{llcccccc}p{3cm}
\begin{tabular*}{\textwidth}{llp{1.8cm}<{\centering}p{1.8cm}<{\centering}p{1.8cm}<{\centering}p{1.8cm}<{\centering}p{1.8cm}<{\centering}p{1.8cm}<{\centering}}
\toprule \hline
 \multirow{2}{*}[-1.0ex]{Dataset}  & \multirow{2}{*}[-1.0ex]{Model}   & \multicolumn{3}{c}{Detection}                                           & \multicolumn{3}{c}{Segmentation (with SAM applied to bboxes)}             \\
                          \cmidrule(lr{0pt}){3-5} \cmidrule(lr{0pt}){6-8} 
                          &                      & $Precision\uparrow$  & $Recall\uparrow$    & $F1\uparrow$          & $MAE\downarrow$      & $F_\beta\uparrow$         & $S_\alpha\uparrow$         \\
                          \midrule
\multirow{4}{*}{DUTS\cite{wang2017learningDUTS}}     & MiniGPT-v2~\cite{chen2023minigptv2}           & 0.296                & \textbf{0.659}       & 0.409                & 0.195                & 0.580                & 0.662                \\
                          &  LLaVA-1.5~\cite{liu2023improvedllava1point5}           & 0.270                & 0.256                & 0.263                & 0.458                & 0.241                & 0.347                \\
                          & Shikra~\cite{chen2023shikra}               & \textbf{0.751}       & 0.583                & \textbf{0.656}       & \textbf{0.102}       & \textbf{0.711}       & \textbf{0.754}       \\
                          & \textcolor{gray}{Upper bound} & \textcolor{gray}{1.000} & \textcolor{gray}{1.000} & \textcolor{gray}{1.000} & \textcolor{gray}{0.054}&\textcolor{gray}{0.905}&\textcolor{gray}{0.892}\\ \hline
\multirow{4}{*}{SOC\cite{fan2018salientSOC}}      & MiniGPT-v2~\cite{chen2023minigptv2}           & 0.289                & \textbf{0.464}       & \textbf{0.359}       & \textbf{0.197}                & \textbf{0.446}                &\textbf{0.578} \\
                          &  LLaVA-1.5~\cite{liu2023improvedllava1point5}           & 0.155                & 0.116                & 0.133                & 0.388                & {0.245}       & 0.314                \\
                          & Shikra~\cite{chen2023shikra}               & \textbf{0.737}       & 0.013                & 0.025                & {0.204}       & {0.245}       & 0.409                \\
                        
                          & \textcolor{gray}{Upper bound} & \textcolor{gray}{1.000} & \textcolor{gray}{1.000} & \textcolor{gray}{1.000} &\textcolor{gray}{0.027}&\textcolor{gray}{0.956}&\textcolor{gray}{0.932}\\ \hline
\multirow{4}{*}{Trans10K\cite{xie2020segmentingTrans10K}} & MiniGPT-v2~\cite{chen2023minigptv2}           & 0.326                & \textbf{0.355}       & 0.340                & 0.185                & 0.624                & 0.656                \\
                          &  LLaVA-1.5~\cite{liu2023improvedllava1point5}           & 0.452                & 0.250                & 0.322                & 0.287                & 0.441                & 0.490                \\
                          & Shikra~\cite{chen2023shikra}               & \textbf{0.614}       & 0.322                & \textbf{0.431}       & \textbf{0.167}       & \textbf{0.692}       & \textbf{0.683}       \\
                          & \textcolor{gray}{Upper bound} & \textcolor{gray}{1.000} & \textcolor{gray}{1.000} & \textcolor{gray}{1.000} &\textcolor{gray}{0.108}&\textcolor{gray}{0.868}&\textcolor{gray}{0.824}\\ \hline
\multirow{4}{*}{COD10K\cite{fan2022concealed}}   & MiniGPT-v2~\cite{chen2023minigptv2}           & \textbf{0.338}       & \textbf{0.575}       & \textbf{0.426}       & 0.308                & 0.390                & 0.524                \\
                          &  LLaVA-1.5~\cite{liu2023improvedllava1point5}           & 0.284                & 0.270                & 0.277                & 0.454                & 0.226                & 0.352                \\
                          & Shikra~\cite{chen2023shikra}               & 0.327                & 0.301                & 0.313                & \textbf{0.166}       & \textbf{0.456}       & \textbf{0.585}       \\ 
                          & \textcolor{gray}{Upper bound} & \textcolor{gray}{1.000} & \textcolor{gray}{1.000} & \textcolor{gray}{1.000} &\textcolor{gray}{0.054}&\textcolor{gray}{0.808}&\textcolor{gray}{0.844}\\ \hline
\bottomrule               
\end{tabular*}
% \vspace{5pt}
\end{table*}

\begin{table*}[t!]
\footnotesize
\centering
\caption{Detection and segmentation results of MiniGPT-v2, LLaVA-1.5, and Shikra in healthcare. The symbols $\uparrow$/$\downarrow$ indicate that a higher/lower score is better, while the highest scores are marked in bold. The \textcolor{gray}{upper bound} (on ground truth bounding boxes) of detection and segmentation via LVLMs in diverse specialized tasks is marked in gray.}
\label{tab:lvlm_dec_and_seg_healthcare}
% \tabcolsep 11pt
% \begin{tabular*}{\textwidth}{cccccccc}
\begin{tabular*}{\textwidth}{llp{1.8cm}<{\centering}p{1.8cm}<{\centering}p{1.8cm}<{\centering}p{1.8cm}<{\centering}p{1.8cm}<{\centering}p{1.8cm}<{\centering}}
\toprule\hline
 \multirow{2}{*}[-1.0ex]{Dataset}  & \multirow{2}{*}[-1.0ex]{Model}   & \multicolumn{3}{c}{Detection}                                      & \multicolumn{3}{c}{Segmentation (with SAM applied to bboxes)}                                    \\
                          \cmidrule(lr{0pt}){3-5} \cmidrule(lr{0pt}){6-8} 
                          &                      & $Precision\uparrow$  & $Recall\uparrow$    & $F1\uparrow$          & $MAE\downarrow$      & $F_\beta\uparrow$         & $S_\alpha\uparrow$         \\
                          \midrule
\multirow{4}{*}{ColonDB\cite{tajbakhsh2015automatedColonDB}}  & MiniGPT-v2~\cite{chen2023minigptv2}           & 0.153                & \textbf{0.287}       & 0.199                & 0.322                & \textbf{0.281}       & 0.467                \\
                          &  LLaVA-1.5~\cite{liu2023improvedllava1point5}           & \textbf{0.245}       & 0.237                & \textbf{0.241}       & \textbf{0.190}       & 0.273                & \textbf{0.504}       \\
                          & Shikra~\cite{chen2023shikra}               & 0.163                & 0.163                & 0.163                & 0.540                & 0.232                & 0.338                \\
                          & \textcolor{gray}{Upper bound} & \textcolor{gray}{1.000} & \textcolor{gray}{1.000} & \textcolor{gray}{1.000} &\textcolor{gray}{0.019}&\textcolor{gray}{0.906}&\textcolor{gray}{0.916}\\  \hline
\multirow{4}{*}{ETIS\cite{silva2014towardETIS}}     & MiniGPT-v2~\cite{chen2023minigptv2}           & 0.116                & \textbf{0.221}       & \textbf{0.152}       & \textbf{0.523}       & 0.196                & \textbf{0.336}       \\
                          &  LLaVA-1.5~\cite{liu2023improvedllava1point5}           & 0.092                & 0.087                & 0.089                & 0.640                & 0.197                & 0.268                \\
                          & Shikra~\cite{chen2023shikra}               & \textbf{0.148}       & 0.139                & 0.144                & 0.675                & \textbf{0.206}       & 0.261                \\
                          & \textcolor{gray}{Upper bound} & \textcolor{gray}{1.000} & \textcolor{gray}{1.000} & \textcolor{gray}{1.000} &\textcolor{gray}{0.006}&\textcolor{gray}{0.912}&\textcolor{gray}{0.947}\\ \hline
\multirow{4}{*}{ISIC\cite{codella2018skinISIC2017}}     & MiniGPT-v2~\cite{chen2023minigptv2}           & 0.321                & \textbf{0.610}       & 0.421                & 0.350                & \textbf{0.561}       & \textbf{0.509}       \\
                          &  LLaVA-1.5~\cite{liu2023improvedllava1point5}           & \textbf{0.573}       & 0.568                & \textbf{0.570}       & 0.404                & 0.519                & 0.442                \\
                          & Shikra~\cite{chen2023shikra}               & 0.398                & 0.398                & 0.398                & \textbf{0.348}       & 0.430                & 0.448                \\
                          % & Baseline             &     -                &  -                   & -                    & 0.106                & 0.842                & 0.765                \\
                          & \textcolor{gray}{Upper bound} & \textcolor{gray}{1.000} & \textcolor{gray}{1.000} & \textcolor{gray}{1.000} &\textcolor{gray}{0.106}&\textcolor{gray}{0.842}&\textcolor{gray}{0.765}\\ \hline
\bottomrule            
\end{tabular*}
% \vspace{4pt}
\end{table*}

\begin{table*}[t!]
\footnotesize
\centering
\caption{Detection and segmentation results of MiniGPT-v2, LLaVA-1.5, and Shikra in industrial scenarios. The symbols $\uparrow$/$\downarrow$ indicate that a higher/lower score is better, while the highest scores are marked in bold. The \textcolor{gray}{upper bound} (on ground truth bounding boxes) of detection and segmentation via LVLMs in diverse specialized tasks is marked in gray.}
\label{tab:lvlm_dec_and_seg_industrial}
% \tabcolsep 11pt
% \begin{tabular*}{\textwidth}{cccccccc}
\begin{tabular*}{\textwidth}{llp{1.8cm}<{\centering}p{1.8cm}<{\centering}p{1.8cm}<{\centering}p{1.8cm}<{\centering}p{1.8cm}<{\centering}p{1.8cm}<{\centering}}
\toprule \hline
\multirow{2}{*}[-1.0ex]{Dataset}  & \multirow{2}{*}[-1.0ex]{Model}   & \multicolumn{3}{c}{Detection}                                      & \multicolumn{3}{c}{Segmentation (with SAM applied to bboxes)}                                    \\
                          \cmidrule(lr{0pt}){3-5} \cmidrule(lr{0pt}){6-8} 
                          &                      & $Precision\uparrow$  & $Recall\uparrow$    & $F1\uparrow$          & $MAE\downarrow$      & $F_\beta\uparrow$         & $S_\alpha\uparrow$         \\
                          \midrule
\multirow{4}{*}{MVTec AD\cite{bergmann2021mvtec}} & MiniGPT-v2~\cite{chen2023minigptv2}           & 0.107                & 0.212                & 0.142                & 0.511                & 0.381                & 0.292                 \\
                          &  LLaVA-1.5~\cite{liu2023improvedllava1point5}           & 0.081                & 0.065                & 0.072                & 0.580                & 0.061                & 0.239                 \\
                          & Shikra~\cite{chen2023shikra}               & \textbf{0.355}       & \textbf{0.281}       & \textbf{0.314}       & \textbf{0.090}       & \textbf{0.425}       & \textbf{0.622}        \\
                          & \textcolor{gray}{Upper bound} & \textcolor{gray}{1.000} & \textcolor{gray}{1.000} & \textcolor{gray}{1.000} &\textcolor{gray}{0.032}&\textcolor{gray}{0.784}&\textcolor{gray}{0.831}\\ \hline
\multirow{4}{*}{VisA\cite{zou2022spotVisA}}     & MiniGPT-v2~\cite{chen2023minigptv2}           & 0.009                & 0.032                & 0.014                & 0.211                & 0.051                & 0.410                 \\
                          &  LLaVA-1.5~\cite{liu2023improvedllava1point5}           & 0.007                & 0.007                & 0.007                & 0.532                & 0.016                & 0.259                 \\
                          & Shikra~\cite{chen2023shikra}               & \textbf{0.107}       & \textbf{0.076}       & \textbf{0.089}       & \textbf{0.100}       & \textbf{0.153}       & \textbf{0.505}        \\
                          & \textcolor{gray}{Upper bound} & \textcolor{gray}{1.000} & \textcolor{gray}{1.000} & \textcolor{gray}{1.000} &\textcolor{gray}{0.004}&\textcolor{gray}{0.697}&\textcolor{gray}{0.819}\\ \hline
\bottomrule
                    
\end{tabular*}
% \vspace{4pt}
\end{table*}

\subsubsection{Benchmark Datasets}
Nine datasets from SOD (DUTS~\cite{wang2017learningDUTS} and SOC ~\cite{fan2018salientSOC}), COD (COD10K~\cite{fan2022concealed}), TOD (Trans10K~\cite{xie2020segmentingTrans10K}), skin lesion detection (ColonDB~\cite{tajbakhsh2015automatedColonDB}), polyp detection (ETIS~\cite{silva2014towardETIS} and  ISIC~\cite{codella2018skinISIC2017}), and AD (MVTec AD~\cite{bergmann2021mvtec} and VisA~\cite{zou2022spotVisA}) mentioned in \S~\ref{sec:cls_dataset} are utilized to evaluate the localization capability.
Since these datasets only provide mask annotations, we derive ground truth bounding boxes from these masks to evaluate the detection performance.
Given the inherent difficulty of LVLMs in judging target existence in negative samples as demonstrated in \S~\ref{sec:perception}, we solely utilize positive samples from the aforementioned datasets to assess the localization capability.

\subsubsection{Result Analyses and Discussions}
The results are reported in Tables~\ref{tab:lvlm_dec_and_seg_natural}-\ref{tab:lvlm_dec_and_seg_industrial}, from which several observations can be derived.

\begin{figure*}[t]
  \centering
  % % \vspace{-5pt}
  \includegraphics[width=\linewidth]{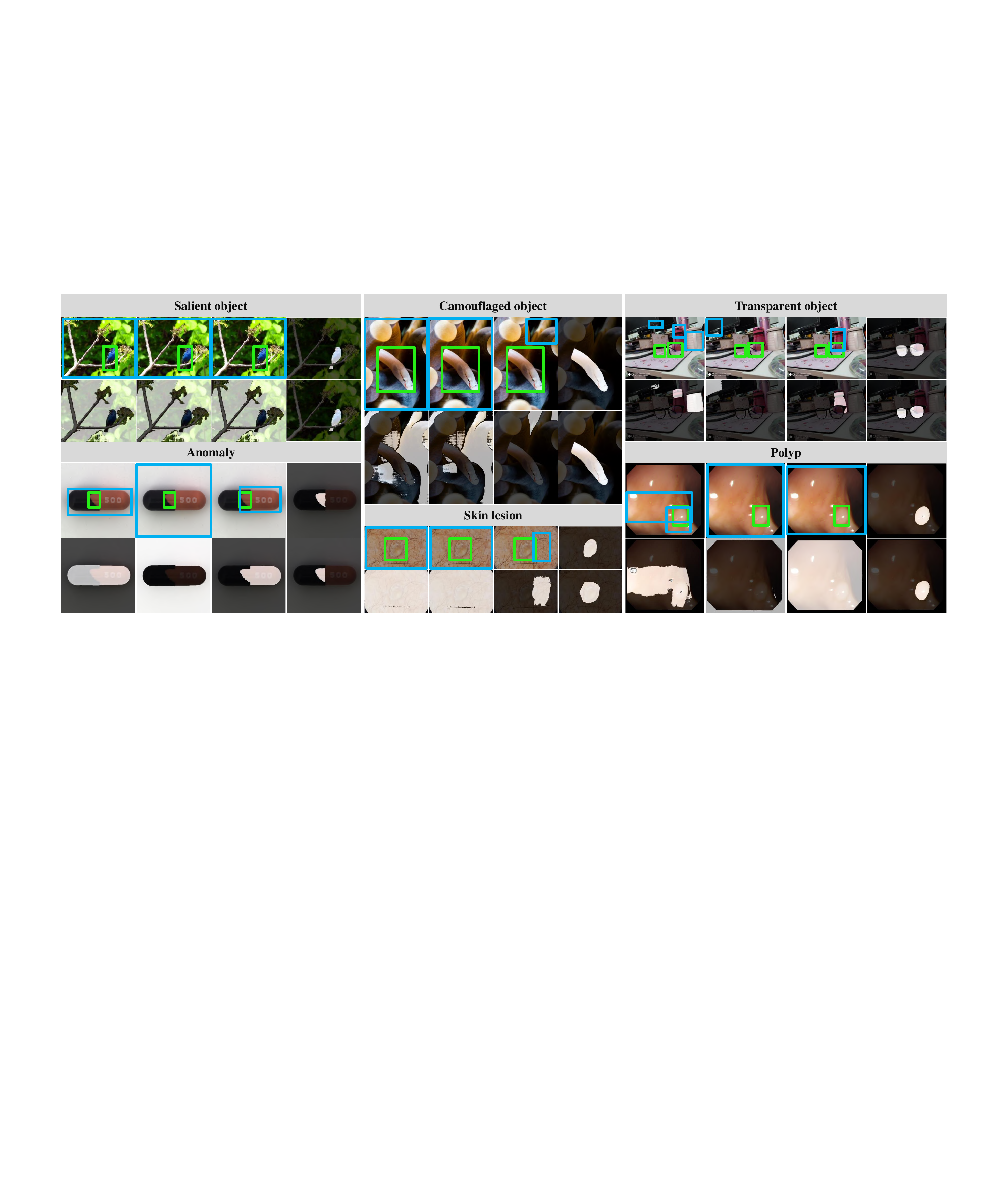}
  % \vspace{-20pt}
  \caption{Detection and segmentation results of three LVLMs in six specialized tasks. The predicted bounding boxes and ground truth are marked with blue and green. From left to right in each scenario: detection (top) and segmentation (bottom) results of MiniGPT-v2~\cite{chen2023minigptv2}, LLaVA-1.5~\cite{liu2023improvedllava1point5}, and Shikra~\cite{chen2023shikra}, as well as segmentation results of upper bound (top) and the ground truth masks (bottom).}
 % \vspace{-10pt}
  \label{fig:quantitative_location}
\end{figure*}

\textbf{Promising yet insufficient localization capability for specific tasks}. 
The results in Tables~\ref{tab:lvlm_dec_and_seg_natural}-\ref{tab:lvlm_dec_and_seg_industrial} show that these LVLMs hold promise for addressing specialized tasks without requiring domain-specific fine-tuning, particularly in natural scenarios. While Shikra and MiniGPT-v2 show better localization capability compared to LLaVA-1.5, superior segmentation performance is achieved by Shikra on DUTS ($S_\alpha$ score 0.754) and Trans10K ($S_\alpha$ score 0.683) when only provided with category names. 
However, their detection and segmentation performance is found inadequate as their performance is much lower than that of the upper bound. This indicates their insufficient localization capability in these specialized tasks. Specifically, the low scores in terms of $\rm Precision$ and $\rm Recall$ demonstrate that these models struggle to generate precise bounding boxes (i.e., most predicted boxes are inaccurate) and identify targets (i.e., most objects are missed for detection). 
These limitations ultimately restrict the final segmentation performance of LVLMs on specialized tasks.

\textbf{Superior performance in natural scenarios}. According to the results presented in Tables~\ref{tab:lvlm_dec_and_seg_natural}-\ref{tab:lvlm_dec_and_seg_industrial}, these models demonstrate superior performance in natural scenarios, especially on DUTS and Trans10K. 
The underlying reason may be that transparent and salient objects are more prevalent and exhibit common attributes. Conversely, medical and abnormal images are relatively scarce and with complex characteristics, thereby posing greater challenges for LVLMs.

Furthermore, we illustrate the detection and segmentation results in Fig.~\ref{fig:quantitative_location}. As evidence, these models face challenges in providing accurate bounding boxes, consequently resulting in subpar segmentation performance. These findings underscore their limited localization capabilities in specialized tasks.

\begin{figure*}[t!]
  \centering
  \vspace{-5pt}
  \includegraphics[width=\linewidth]{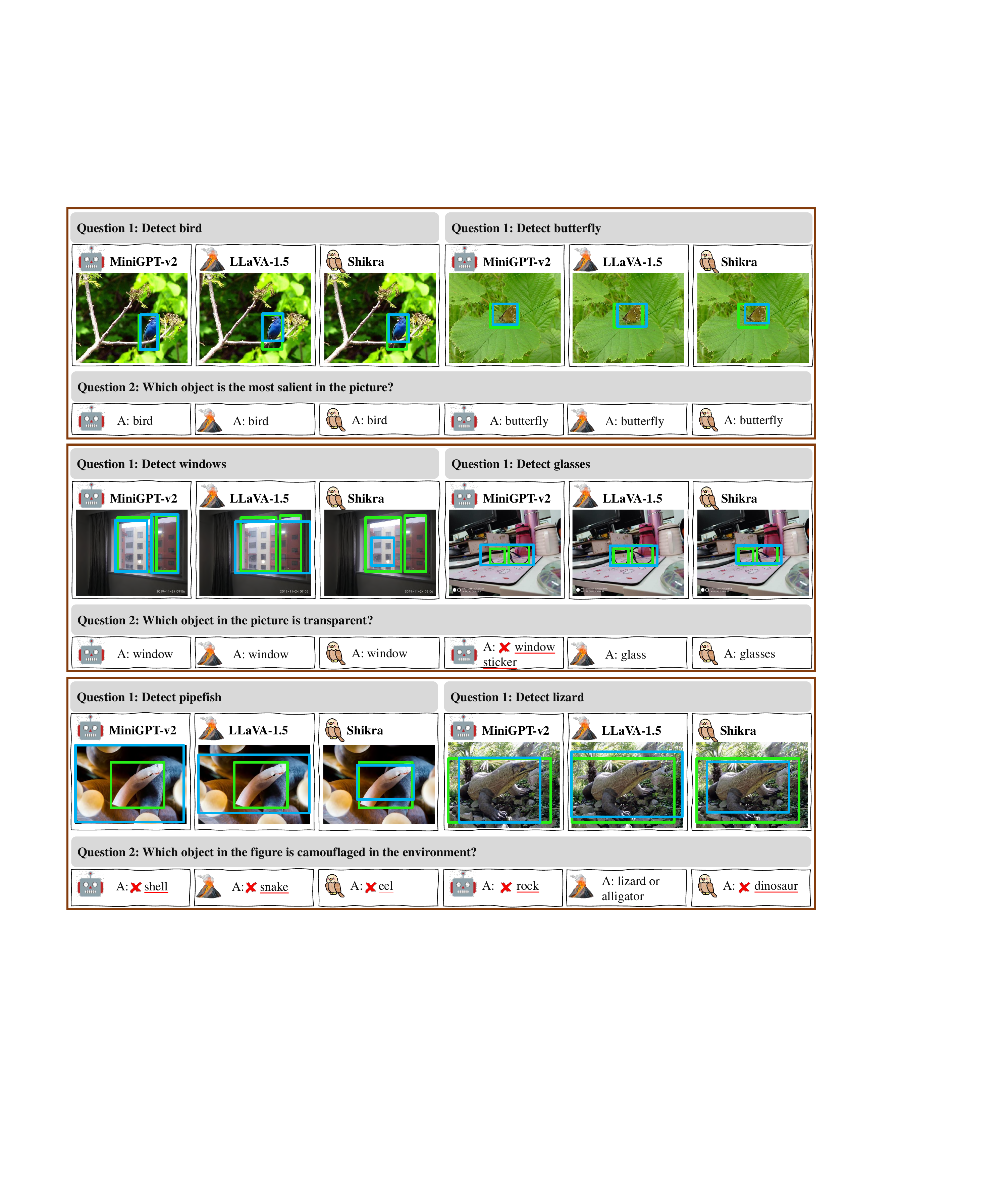}
  \vspace{-13pt}
  \caption{Responses of three LVLMs regarding locating given objects and recognizing objects of specific types. Predicted bounding boxes and ground truth are marked in blue and green. From top to bottom: examples of salient object detection, transparent object detection, and camouflaged object detection. Incorrect responses are marked with red underlines and crosses.}
 \vspace{-10pt}
  \label{fig:qualitative_natural}
\end{figure*}

\begin{figure*}[ht!]
  \centering
  \vspace{-2pt}
  \includegraphics[width=\linewidth]{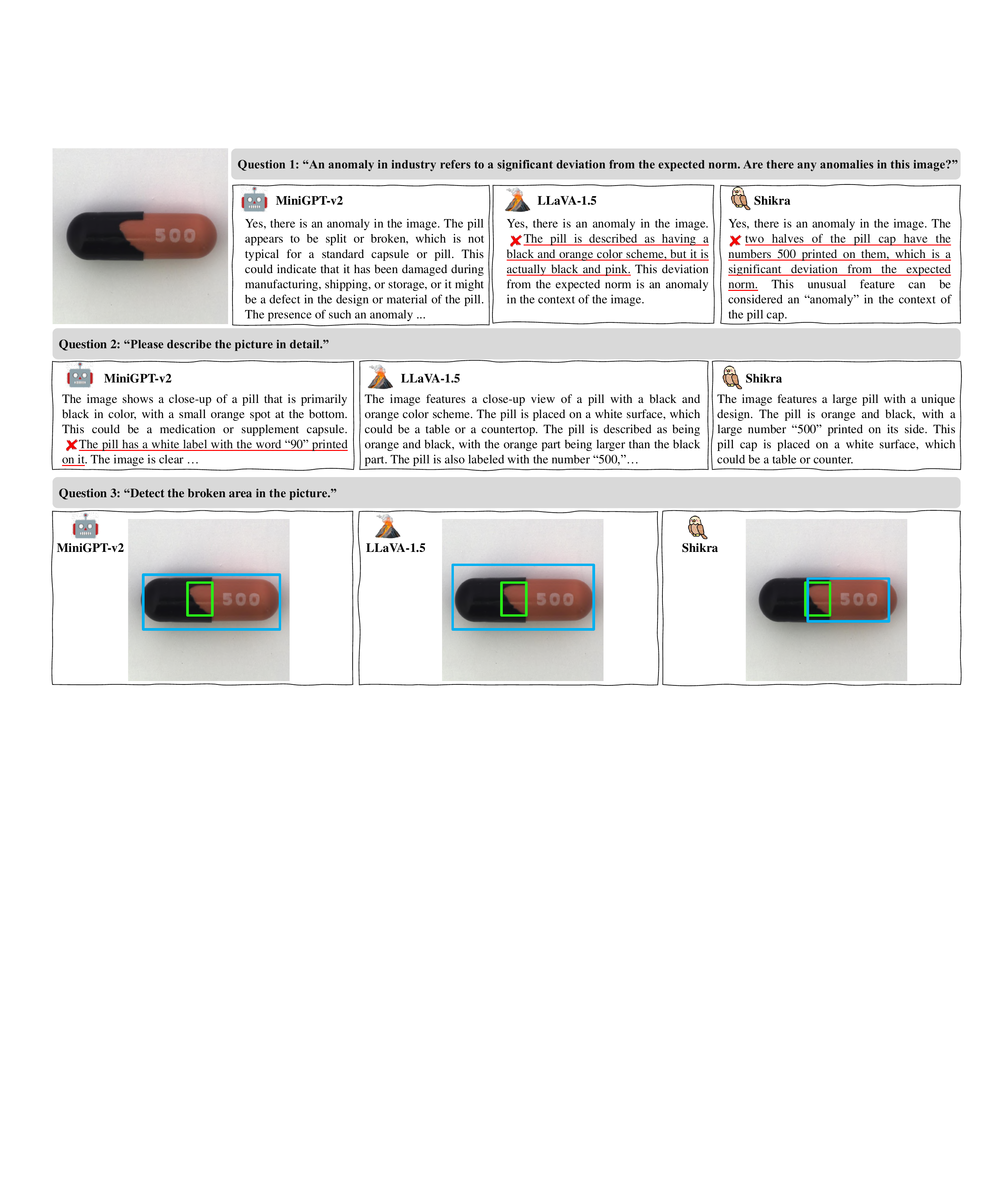}
  \vspace{-16pt}
  \caption{Responses of three LVLMs regarding recognizing and locating the anomaly. Predicted bounding boxes and ground truth are marked in blue and green, respectively. The incorrect responses are marked with red underlines and crosses.}
 \vspace{-5pt}
  \label{fig:qualitative_industry}
\end{figure*}

% \subsubsection{Locating specific object}\label{sec:quali_det}

\subsection{Uncovering Insights into Failure Cases}\label{sec:quali_det}
\begin{figure*}[!ht]
\vspace{-2pt}
\centering
\includegraphics[width=\linewidth]{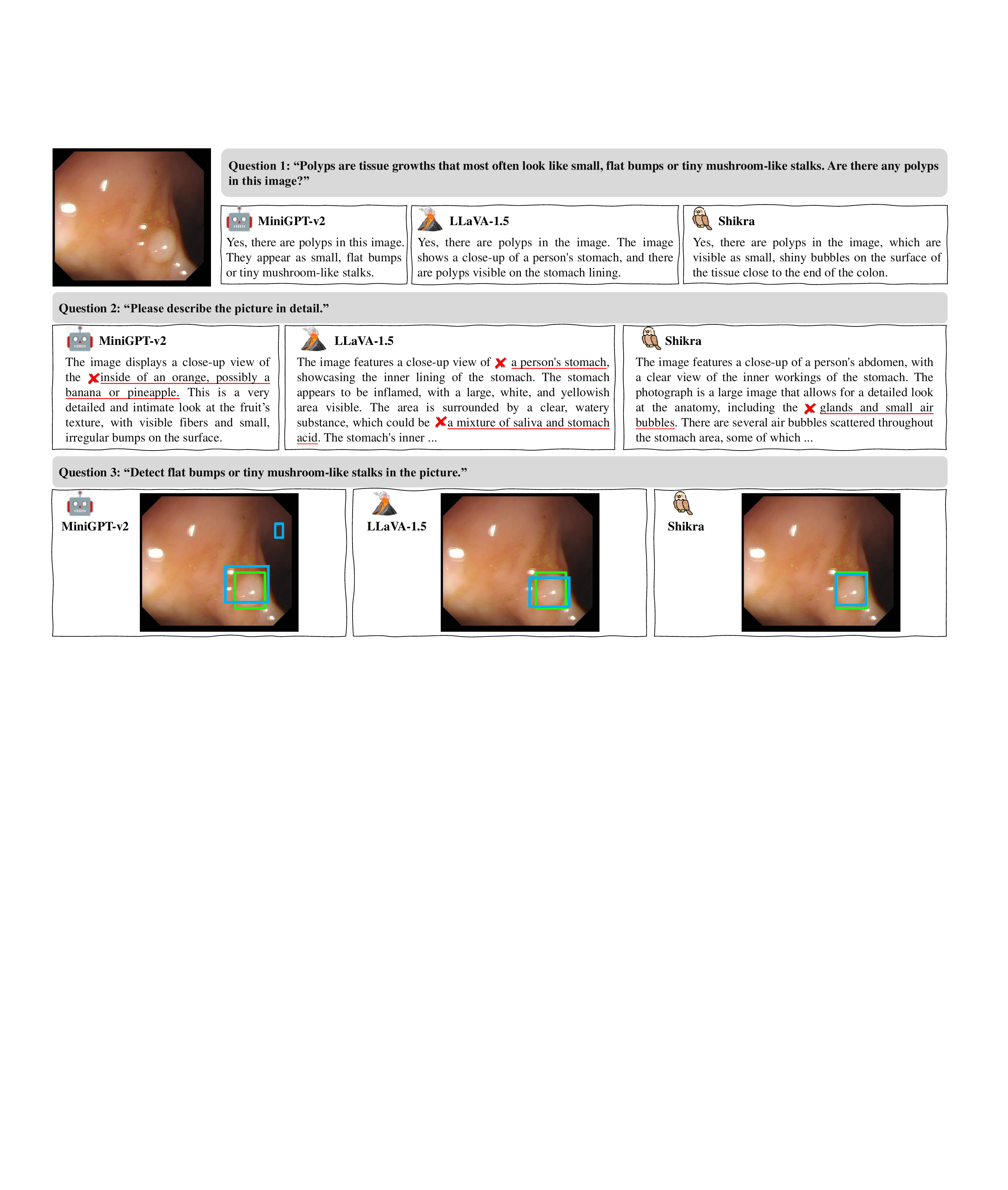}
\vspace{-16pt}
  \caption{Responses of three LVLMs regarding recognizing and locating the colon polyp. Predicted bounding boxes and ground truth are marked in blue and green, respectively. Incorrect responses are marked with red underlines and crosses.}
 \vspace{-5pt}
  \label{fig:qualitative_healthcare}

\end{figure*}

As mentioned in \S~\ref{sec:quanti_det}, we evaluate the localization capability of LVLMs by solely specifying object types.
This setting concurrently evaluates their recognition, reasoning, and localization capabilities by requiring models to accurately perceive each object.
Therefore, we sought to gain insight into the underlying reasons behind such inability by breaking down the question in \S~\ref{sec:quanti_det} into multiple questions. We focus on failure cases of LVLMs and prompt them with multiple questions.  
In natural scenarios, two questions are prompted to assess the models in accurately localizing given objects (``Question 1'') and determining the target of specific types (``Question 2''). 
In industrial scenarios, because anomalies usually are hard to tell their detailed categories,
we evaluate the recognition of anomalies by querying the existence (``Question 1'') and image description (``Question 2''), and further test their capability to locate anomalous areas by providing corresponding descriptions (``Question 3''). 
In healthcare (colon polyp detection), we follow the same protocol as in industrial cases. 
The results are separately presented in Figs.~\ref{fig:qualitative_natural}-\ref{fig:qualitative_healthcare}. 
Two underlying reasons for failing to locate can then be drawn.

\textbf{Decreased robustness in complex problems}.
The results in Fig.~\ref{fig:qualitative_natural} reveal that these models are good at locating a given object or inferring the target, especially for salient and transparent objects.  
However, they make errors when asked to locate the target types directly, as shown in Fig.~\ref{fig:quantitative_location}. This failure indicates that they exhibit decreased robustness or are unskilled when faced with more complex and abstracted problems. That is, they need to understand the notion of complex concept of ``salient''.
It is worth noting that LVLMs achieve lower performance on camouflaged objects, which could be attributed to the resemblance of camouflaged objects to their surroundings. Such failures also demonstrate the challenge faced by LVLMs in accurately categorizing these objects, as mentioned in \S~\ref{sec:quali_cls}.

\textbf{Limited cognition toward medical images and anomalies}. 
Fig.~\ref{fig:qualitative_industry} and Fig.~\ref{fig:qualitative_healthcare} clearly demonstrate the limited cognition of LVLMs on medical images and anomalies. For instance, LLaVA-1.5 and Shikra erroneously categorize the ``black and orange color scheme'' and ``the number 500'' as anomalies (as shown in Fig.~\ref{fig:qualitative_industry}), while MiniGPT-v2 incorrectly recognizes colon image as ``the inside of an orange'' (as shown in Fig.~\ref{fig:qualitative_healthcare}). 
Despite their limitations, these LVLMs show superior localization capabilities on polyp when provided with relevant descriptions (as evidenced by the responses to ``Question 3'' in Fig.~\ref{fig:qualitative_healthcare}). Nevertheless, there still remains room for enhancement in localization regarding anomaly detection.

\begin{table*}[t!]
\footnotesize
\centering
\vspace{-5pt}
\caption{Performance summary of MiniGPT-v2, LLaVA-1.5, and Shikra in SOD, TOD, COD, polyp detection (PD), skin lesion detection (SLD), and AD. Thresholds are established at 60\% and 80\% of the upper-bound performance to categorize model performance into three intuitive levels: low (\textcolor{myGreen}{\textbf{L}}), medium (\textcolor{blue}{\textbf{M}}), and high (\textcolor{red}{\textbf{H}}). The notation ``-'' denotes inconclusive cases, since the evaluation is performed only on the positive sets, while the models incur the over-positive issue.}
\vspace{-5pt}
\label{tab:lvlm_summary}
\begin{tabular}{lcccccccccccc}
\toprule \hline
    \multirow{2}{*}[-2.5ex]{Model} & \multicolumn{6}{c}{Recognition}                      & \multicolumn{6}{c}{Localization}                      \\ \cmidrule(lr{0pt}){2-7} \cmidrule(lr{0pt}){8-13} 
     & \multicolumn{3}{c}{Natural}                      & \multicolumn{2}{c}{Healthcare}    & \multicolumn{1}{c}{Industrial }   & \multicolumn{3}{c}{Natural}                      & \multicolumn{2}{c}{Healthcare}    & \multicolumn{1}{c}{Industrial }                \\ \cmidrule(lr{0pt}){2-4} \cmidrule(lr{0pt}){5-6} \cmidrule(lr{0pt}){7-7} \cmidrule(lr{0pt}){8-10} \cmidrule(lr{0pt}){11-12}\cmidrule(lr{0pt}){13-13}
           & SOD & TOD & COD & PD & SLD & AD & SOD & TOD & COD & PD & SLD & AD \\ \hline
MiniGPT-v2~\cite{chen2023minigptv2} & \textcolor{myGreen}{\textbf{L}}   & -   & \textcolor{myGreen}{\textbf{L}}   & \textcolor{myGreen}{\textbf{L}}     & -           & \textcolor{blue}{\textbf{M}}  & \textcolor{blue}{\textbf{M}}   & \textcolor{blue}{\textbf{M}}   & \textcolor{blue}{\textbf{M}}   & \textcolor{myGreen}{\textbf{L}}     & \textcolor{blue}{\textbf{M}}           & \textcolor{myGreen}{\textbf{L}}  \\
LLaVA-1.5~\cite{liu2023improvedllava1point5}  & \textcolor{blue}{\textbf{M}}   & -   & \textcolor{blue}{\textbf{M}}   & \textcolor{myGreen}{\textbf{L}}     & -           & \textcolor{blue}{\textbf{M}}  & \textcolor{myGreen}{\textbf{L}}   & \textcolor{myGreen}{\textbf{L}}   & \textcolor{myGreen}{\textbf{L}}   & \textcolor{myGreen}{\textbf{L}}     & \textcolor{myGreen}{\textbf{L}}           & \textcolor{myGreen}{\textbf{L}}  \\
Shikra~\cite{chen2023shikra}     & \textcolor{myGreen}{\textbf{L}}   & -   & \textcolor{myGreen}{\textbf{L}}   & \textcolor{myGreen}{\textbf{L}}     & -           & \textcolor{blue}{\textbf{M}}  & \textcolor{blue}{\textbf{M}}   & \textcolor{red}{\textbf{H}}   & \textcolor{blue}{\textbf{M}}   & \textcolor{myGreen}{\textbf{L}}     & \textcolor{myGreen}{\textbf{L}}           & \textcolor{blue}{\textbf{M}}  \\ \hline
\bottomrule
\end{tabular}
\vspace{-10pt}
\end{table*}

\begin{figure*}[t]
  \centering
  \vspace{-5pt}
  \includegraphics[width=\linewidth]{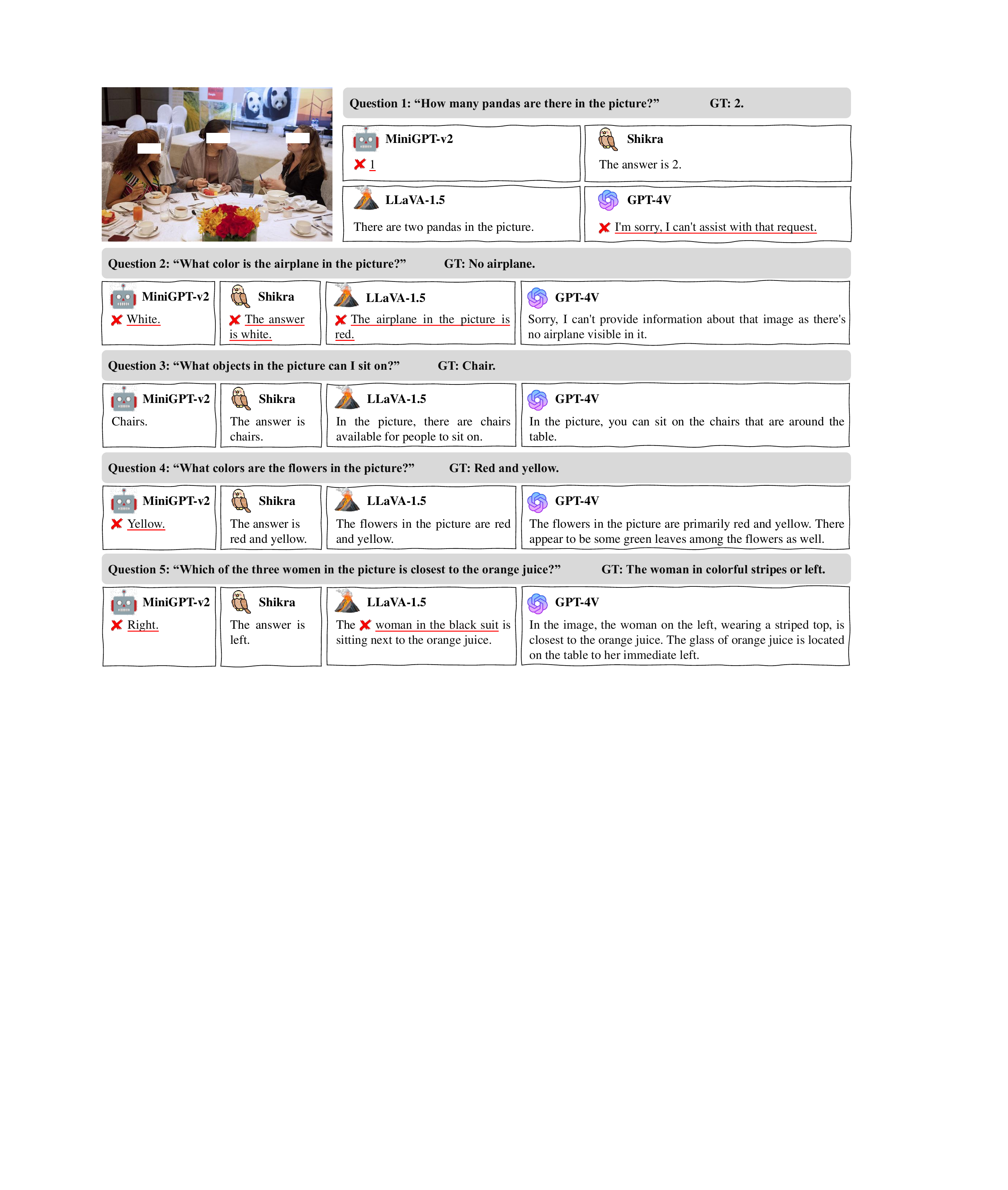}
  \vspace{-20pt}
 \caption{Responses of four LVLMs (MiniGPT-v2, LAVa-1.5, Shikra, and GPT-4V) regarding five general tasks, encompassing object counting (question 1), absurd question answering (question 2), affordance reasoning (question 3), attribute recognition (question 4), and spatial relation reasoning (question 5). Incorrect responses are marked with red underlines and crosses.
 }
 \vspace{-10pt}
  \label{fig:qualitative_multi_questions1}
\end{figure*}

\subsection{Summary}
Sect. 3 evaluates the effectiveness of MiniGPT-v2~\cite{chen2023minigptv2}, LLaVA-1.5~\cite{liu2023improvedllava1point5}, and Shikra~\cite{chen2023shikra} in localizing targets in diverse specialized tasks.
The results reveal that these models hold promise for addressing specialized tasks (particularly in natural scenarios), while Shikra and MiniGPT-v2 show superior localization capability compared to LLaVA-1.5.
Nonetheless, despite the successes, the detection and segmentation performance of these models are still inadequate, indicating a weakness in localization capability for specialized tasks. The limited cognition of medical images and anomalies hampers the \jy{transfer} capability of these LVLMs, whereas decreased robustness when facing complex problems may also be an additional constraint.

\jy{As a summary, we give the general performance of those three models on the six tasks in Table~\ref{tab:lvlm_summary}, where intuitive thresholds are set to categorize the models' average performance into three levels. It is evident that the recognition and localization performance of these models in the six tasks remain insufficient, with most cases exhibiting low (L) or medium (M) performance, indicating less usability in real-world scenarios. Notably, Shikra stands out with a high (H) score on the TOD task, whereas among these models, LLaVA-1.5 demonstrates superiority on recognition compared to MiniGPT-v2 and Shikra. However, the opposite appears to be true for localization.
}

\begin{figure*}[!ht]
  \centering
  \vspace{-5pt}
  \includegraphics[width=\linewidth]{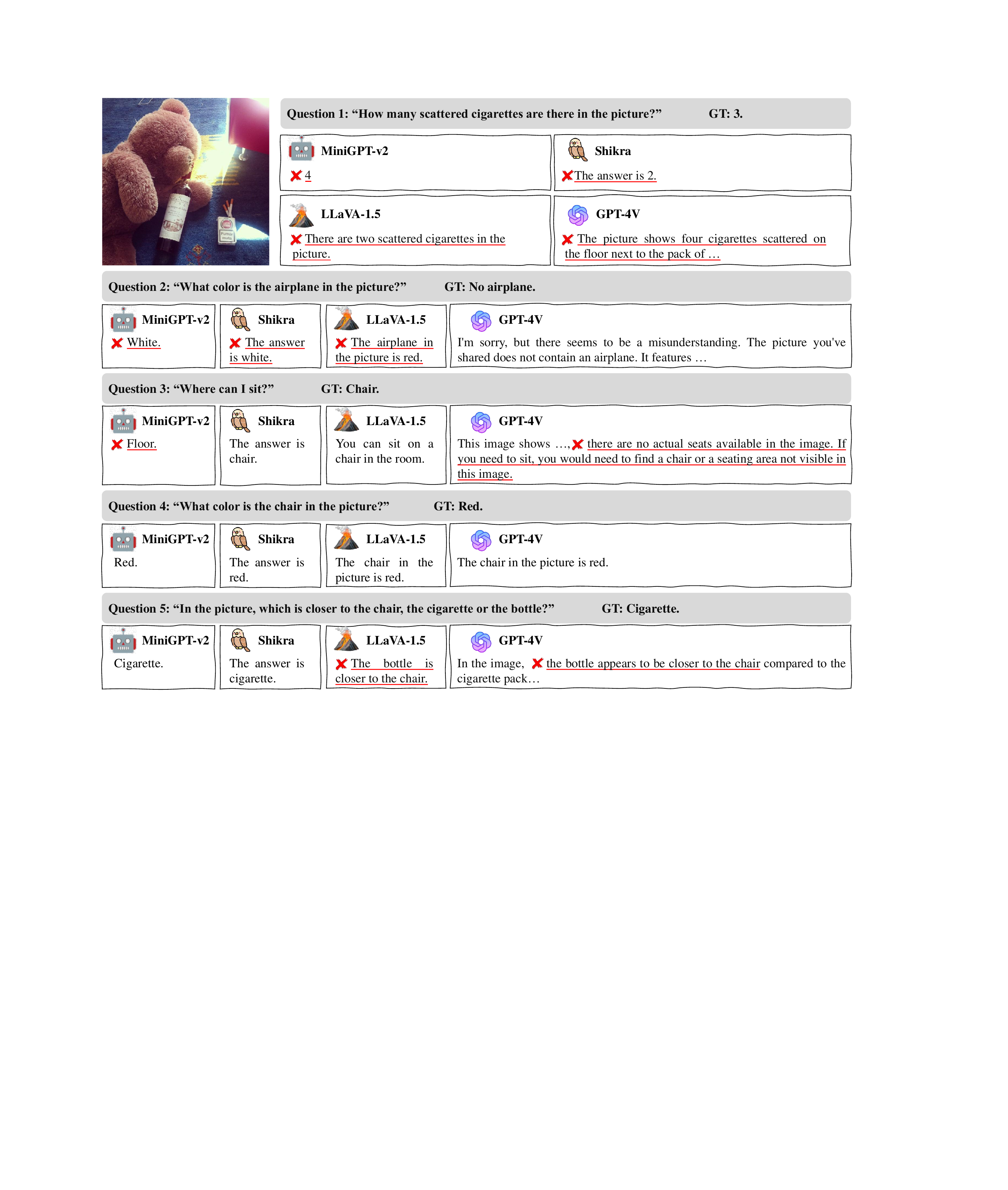}
  \vspace{-20pt}
  \caption{Responses of four LVLMs (MiniGPT-v2, LAVa-1.5, Shikra, and GPT-4V) regarding five general tasks, encompassing object counting (question 1), absurd question answering (question 2), affordance reasoning (question 3), attribute recognition (question 4), and spatial relation reasoning (question 5). Incorrect responses are marked with red underlines and crosses.
 }
 \vspace{-10pt}
  \label{fig:qualitative_multi_questions2}
\end{figure*}

\begin{figure*}[!ht]
  \centering
  \vspace{-5pt}
  \includegraphics[width=\linewidth]{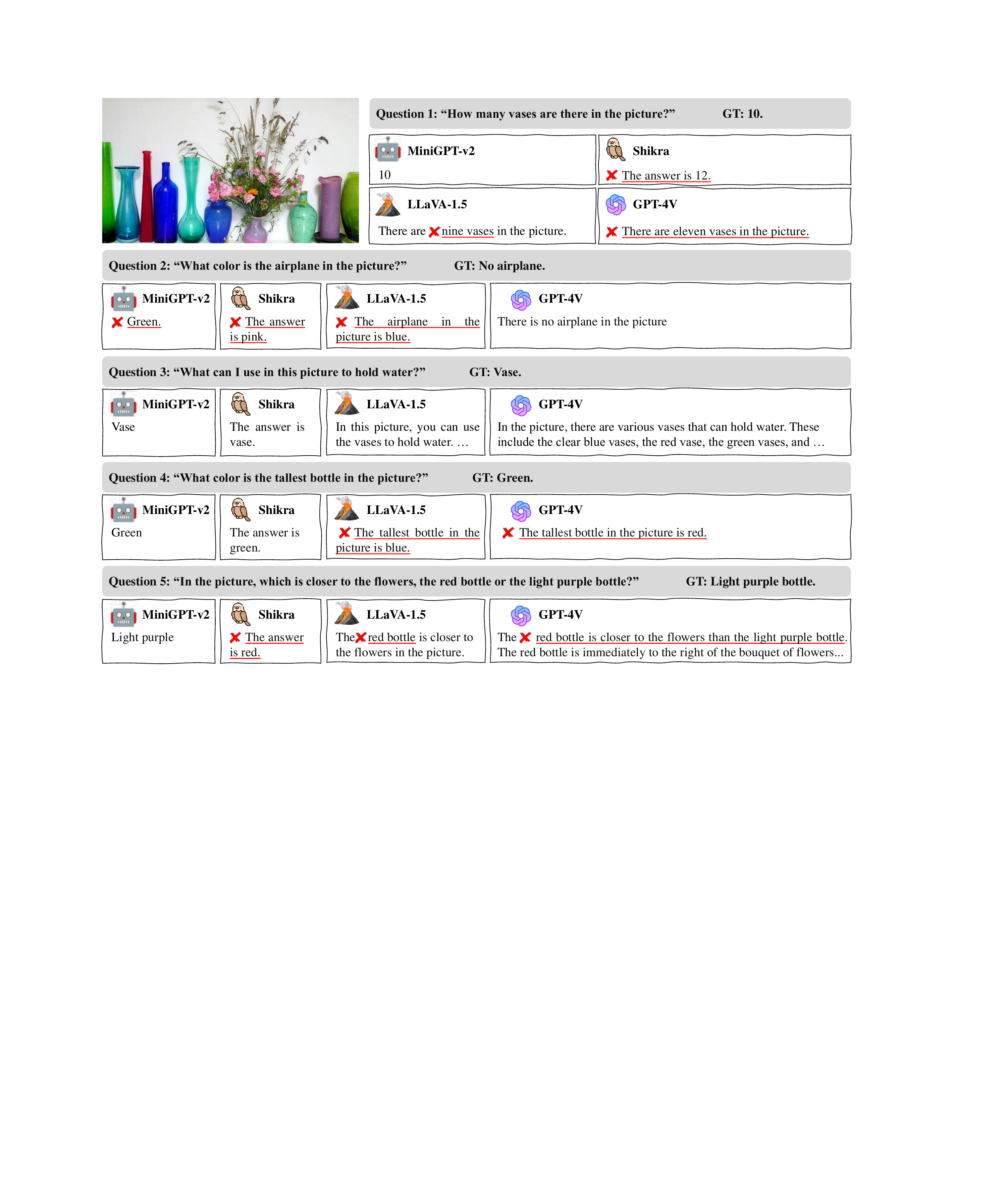}
  \vspace{-20pt}
  \caption{Responses of four LVLMs (MiniGPT-v2, LAVa-1.5, Shikra, and GPT-4V) regarding five general tasks, encompassing object counting (question 1), absurd question answering (question 2), affordance reasoning (question 3), attribute recognition (question 4), and spatial relation reasoning (question 5). Incorrect responses are marked with red underlines and crosses.
 }
 \vspace{-10pt}
  \label{fig:qualitative_multi_questions3}
\end{figure*}

% \subsection{Other capability of LVLM}\label{sec:quali_other}
\section{Capabilities of LVLMs in General Tasks}\label{sec:quali_other}
In this section, we conduct empirical investigations to evaluate the performance of MiniGPT-v2~\cite{chen2023minigptv2}, LLaVA-1.5~\cite{liu2023improvedllava1point5}, Shikra~\cite{chen2023shikra}, and GPT-4V~\cite{openai2023gpt4v} in a diverse range of general tasks.
Given that the recognition and localization of general objects are targets learned by many current LVLMs, and their performance on these tasks has been extensively studied~\cite{chen2023minigptv2,chen2023shikra,liu2023improvedllava1point5}, we shift our focus to five other widely recognized general tasks, including object counting, absurd question answering, affordance reasoning, attribute recognition, and spatial relation reasoning.
We conduct some evaluations of the aforementioned tasks utilizing the COCO~\cite{lin2014microsoftCOCO} dataset and select three representative examples that demonstrate outcomes similar to those of other tests for display, as illustrated in Figs.~\ref{fig:qualitative_multi_questions1}-\ref{fig:qualitative_multi_questions3}.  
Note that, since there are no ground truth annotations/labels regarding the above general tasks in the COCO dataset, only empirical investigations are considered for this evaluation.

\subsection{Object Counting}
Object counting capability serves as a comprehensive indicator of the perception abilities of LVLMs, necessitating not only the recognition of individual targets but also robust counting capabilities.
To evaluate this capability, we prompt LVLMs with questions like ``How many...'' on three images, as shown in Figs.~\ref{fig:qualitative_multi_questions1}-\ref{fig:qualitative_multi_questions3}. The results show that MiniGPT-v2, LLaVA-1.5, and Shikra achieve only one-third accuracy on this evaluation, whereas GPT-4V fails on all tests. This indicates significant room for enhancement in the object counting capability of LVLMs.
Moreover, the inefficacy of these models in counting challenging objects, including small objects (Fig.~\ref{fig:qualitative_multi_questions2}), underscores the importance of enhancing the visual perception capabilities inherent in vision models.

\subsection{Absurd Question Answering}
Recent LVLMs seamlessly integrate textual and visual inputs, achieving superior multi-modal understanding capabilities.
However, an intriguing question arises: what transpires when there is a lack of relevance between text content and images? To explore this, we endeavor to subject these models to absurd questions.
As illustrated in Figs.~\ref{fig:qualitative_multi_questions1}-\ref{fig:qualitative_multi_questions3}, we ask LVLMs ``What color is the airplane in the picture?'' on three different images where no airplane is present.
The results show that while GPT-4V responds with ``no airplane'' on all tests, the other three models always give colors of the nonexistent airplane.
The incorrect responses indicate that in such cases, these models cannot effectively utilize visual information and heavily rely on language input to generate responses.
A potential reason for this phenomenon could be that the textual inputs provide prior information to models, which leads to erroneous judgments of LVLMs~\cite{cui2023holistic}.

\subsection{Affordance Reasoning}
Affordance delineates the cognitive capability of a model regarding the potential functionalities or interactions that an object can offer~\cite{qin2023goodbard}. We delve into affordance reasoning of LVLMs by employing inquiries such as ``What objects in the picture can I ...''.
The outcomes show that these models can accurately identify objects capable of executing the prescribed actions in most cases. 
It is noteworthy that despite the incorrect responses of MiniGPT-v2 and GPT-4V in Fig.~\ref{fig:qualitative_multi_questions2}, which may be caused by the partial visibility of the chair, the mention of reasonable objects demonstrates their ability to establish a connection between behavior and its corresponding object. These results demonstrate their proficient performance in affordance reasoning.

\subsection{Attribute Recognition}
We proceed to validate the object attribute recognition capabilities of the aforementioned models using ``question 4'' with increasing complexity, as illustrated in Figs.~\ref{fig:qualitative_multi_questions1}-\ref{fig:qualitative_multi_questions3}.
From the results, it is clear that there is a greater need for improvement in MiniGPT-v2 compared to the other models, as MiniGPT-v2 shows a deficiency in accurately identifying all the colors of flowers in Fig.~\ref{fig:qualitative_multi_questions1}, while other models demonstrate commendable performance in simple cases (in Fig.~\ref{fig:qualitative_multi_questions1} and Fig.~\ref{fig:qualitative_multi_questions2}). 
Besides, the failures of LLaVA-1.5 and GPT-4V on complex cases (in Fig.~\ref{fig:qualitative_multi_questions3}) indicate that their ability to solve complex problems needs to be further improved.

\subsection{Spatial Relation Reasoning}
We evaluate the spatial relation reasoning capability of LVLMs with the last question in Figs.~\ref{fig:qualitative_multi_questions1}-\ref{fig:qualitative_multi_questions3}.
From this evaluation, we can find that MiniGPT-v2 and Shikra exhibit superior performance by providing incorrect answers only once, while LLaVA-1.5 demonstrates the poorest performance with incorrect answers in all tests. 
The experimental results that compared with MiniGPT-v2 and Shikra, LLaVA-1.5 still has a lot of room for improvement in spatial relation reasoning.

\section{Conclusion}\label{sec:conclusion}

\subsection{Concluding Remarks}
In this study, we assess the progress of LVLMs by evaluating their effectiveness in specialized and general tasks. We begin by evaluating the performance of three recent open-source LVLMs, namely MiniGPT-v2, LLaVA-1.5, and Shikra, in six specialized tasks. These tasks include salient/camouflaged/transparent object detection, polyp detection, skin lesion detection, and industrial anomaly detection.
Additional empirical investigations are conducted on GPT-4V and the aforementioned models to assess their capabilities in general tasks.
The quantitative results indicate that while these models demonstrate promise in specialized tasks, they manifest inadequate \jy{transfer capability} when directly applied to these tasks (as revealed by Table \ref{tab:lvlm_summary}). This limitation stems from their limited understanding of specialized task domains.
In addition to the aforementioned limitation, performance challenges are exacerbated by typical weaknesses of LVLMs, including object hallucination, text-to-image interference, and decreased robustness when confronted with complex problems/concepts.
In addition to the lack of \jy{transfer capability} in specialized tasks, they exhibit suboptimal performance in some general tasks, \ie~ object counting, spatial relation reasoning, and absurd question answering.
The inadequacies observed in both specialized and general tasks highlight a significant gap that LVLMs have yet to bridge on the path toward achieving AGI.
\jy{These challenges also highlight the limitations of LVLMs for real-world applications, particularly in critical domains such as healthcare and industry where errors often yield significant negative consequences. The performance and reliability of LVLMs still fall short of the practical scenarios.
}

\subsection{Discussions} 
Based on the findings presented, we initiate several discussions concerning the application of LVLMs in specialized tasks and their future development. 
We hope that our discussions will stimulate thought and facilitate further exploration in this area.

\textbf{Exploring more effective prompts}. 
Though suboptimal performance has been achieved by current LVLMs, they hold great promise in specialized tasks. Hence, exploring effective strategies to enhance their performance is important, which would benefit both the field of specialized tasks and LVLMs. In this regard, providing additional information within prompts, a practice known as prompt engineering~\cite{gu2023systematicpromptengineering}, is a viable strategy to improve their performance, as demonstrated in Fig.~\ref{fig:qualitative_healthcare}. This strategy has also been verified by some recent studies, which offer more anomaly definitions in prompts~\cite{zhang2023exploringAD} or incorporating additional features of camouflaged targets into the prompts~\cite{tang2023generalizationcod}.

\textbf{Optimizing LVLMs toward specialized tasks}.
As previously noted, prompt engineering has shown promise in improving the performance of LVLMs. 
However, the effectiveness of prompt engineering is still limited when the targets are difficult to be clearly described, such as on COD and AD.
Hence, one of the future research directions involves optimizing LVLMs for specific tasks. This can be achieved by incorporating domain-specific knowledge through techniques such as prompt-tuning or fine-tuning~\cite{gu2023anomalygpt,li2024llavamed,liu2023promoting}, thereby enhancing their performance in specialized tasks.

\textbf{Mitigating hallucination and also other issues}. 
Current LVLMs encounter significant challenges in hallucination~\cite{li2023evaluatingobjecthallucination,cui2023holistic,zhou2023analyzingobjecthallucination,qian2024easyhallucination}, which impact their effectiveness in both general and specific tasks. 
In future research, overcoming these challenges by leveraging advanced techniques, such as hallucination revisor~\cite{zhou2023analyzingobjecthallucination} and chain of visual perception~\cite{tang2023generalizationcod}, 
holds promise for enhancing the effectiveness of LVLMs in diverse tasks and facilitating broader application of these models.
Besides, it is equally imperative to implement suitable strategies, such as data augmentation that eliminate co-occurrence patterns~\cite{kim2023exposinghallucination} to address the issues. 
Apart from the hallucination, these models encounter additional challenges, including decreased robustness when confronted with complex problems and reduced effectiveness in numerous general tasks, underscoring the fact that the comprehensive capabilities of current LVLMs remain limited. 
Future research is anticipated to leverage increasingly challenging datasets/problems while also providing detailed and specific procedures in instruction tuning~\cite{chen2023shikra,wu2023improvingCrossTaskGeneralization} to enhance the comprehensive capabilities of LVLMs. Besides, adopting advanced techniques such as feedback/reward mechanisms~\cite{chen2023feedback,yan2024vigor} and integrating expert models~\cite{jiao2024enhancing} are also viable avenues to enhance their capabilities.

\jy{
\textbf{Incorporating additional visual information}.
Current LVLMs exhibit a significant limitation in leveraging visual information, as they are restricted to utilizing a single image, typically an RGB image, for each task~\cite{yao2023deepspeed}.
It is widely recognized that for certain visual tasks, such as object detection and recognition in complex scenes (\eg, those with heavy background clutter), relying solely on a single modality of visual information poses significant challenges~\cite{fu2021siamese,fu2022light}. Therefore, the visual perceptual capabilities of LVLMs will be greatly limited when applied to these tasks.
To address this issue, a potential avenue for the future development of LVLMs is to integrate complementary visual information, such as depth~\cite{he2022fewshot,zhou2021specificity,chen2021rgb,fu2019deepside,zhang2021depth} and focus cues~\cite{fu2022light} to augment their perceptual capabilities, the effectiveness of which has been extensively validated in the domain of computer vision.
}

\jy{
\textbf{Other potential applications of LVLMs}.
Despite the existing room for improvement, LVLMs have exhibited remarkable proficiency in tasks such as image summarization/description and visual question answering. Their superior proficiency in these fundamental tasks holds promise for their application in diverse domains. For example, harnessing the aforementioned capabilities of LVLMs to assist data annotation can significantly reduce annotation cost, which can further provide more support for training expert models or enhancing model capabilities~\cite{zhong2024vlm}. 
Besides, the potential of LVLMs in effectively performing a wide range of video-language tasks, such as video retrieval and video description, has been demonstrated remarkably~\cite{wang2022language}. Inspired by this, LVLMs can be further applied to address other video-language tasks, such as video object segmentation~\cite{he2024decouplingDshmp,ding2023mevisMeVis,ding2023moseMoSE} and video captioning~\cite{zhang2019reconstruct} by first generating object descriptions and then conducting the tasks in a single frame.
}

\ifCLASSOPTIONcaptionsoff
  \newpage
\fi

% {
% \bibliographystyle{IEEEtran}
% \bibliography{bibliography}

\begin{thebibliography}{10}

\bibitem{openai2023gpt3}
Brown, T., Mann, B., Ryder, N., Subbiah, M., Kaplan, J.~D., Dhariwal, P., et~al. (2020).
\newblock Language models are few-shot learners.
\newblock In Larochelle, H., Ranzato, M., Hadsell, R., et~al. (Eds.), {\em
  Proceedings of the 34th international conference on neural information
  processing systems}, pages 1877--1901. Red Hook: Curran Associates.

\bibitem{touvron2023llama}
Touvron, H., Lavril, T., Izacard, G., Martinet, X., Lachaux, M.-A., Lacroix, T., et~al. (2023).
\newblock Llama: Open and efficient foundation language models.
\newblock {\em arXiv preprint arXiv:2302.13971}.

\bibitem{liu2023visualllava}
Liu, H., Li, C., Wu, Q., Lee, Y.~J. (2023b).
\newblock Visual instruction tuning.
\newblock In OH, A., Naumann, T., Globerson, A., et~al. (Eds.), {\em
  Proceedings of the 37th international conference on neural information
  processing systems}. Red Hook: Curran Associates.

\bibitem{chen2023minigptv2}
Chen, J., Zhu, D., Shen, X., Li, X., Liu, Z., Zhang, P., et~al. (2023b).
\newblock Minigpt-v2: large language model as a unified interface for
  vision-language multi-task learning.
\newblock {\em arXiv preprint arXiv:2310.09478}.

\bibitem{openai2023gpt4v}
Achiam, J., Adler, S., Agarwal, S., Ahmad, L., Akkaya, I., Aleman, F.~L., et~al. (2023).
\newblock Gpt-4 technical report.
\newblock {\em arXiv preprint arXiv:2303.08774}.

\bibitem{fu2023challengerGemini}
Fu, C., Zhang, R., Lin, H., Wang, Z., Gao, T., Luo, Y., et~al. (2023b).
\newblock A challenger to gpt-4v? early explorations of gemini in visual
  expertise.
\newblock {\em arXiv preprint arXiv:2312.12436}.

\bibitem{qin2023goodbard}
Qin, H., Ji, G.-P., Khan, S., Fan, D.-P., Khan, F.~S., Gool, L.~V. (2023).
\newblock How good is google bard’s visual understanding? an empirical study
  on open challenges.
\newblock {\em Machine Intelligence Research}, 20:605--613.

\bibitem{liu2023improvedllava1point5}
Liu, H., Li, C., Li, Y., Lee, Y.~J. (2023a).
\newblock Improved baselines with visual instruction tuning.
\newblock {\em arXiv preprint arXiv:2310.03744}.

\bibitem{chen2023shikra}
Chen, K., Zhang, Z., Zeng, W., Zhang, R., Zhu, F., Zhao, R. (2023c).
\newblock Shikra: Unleashing multimodal llm's referential dialogue magic.
\newblock {\em arXiv preprint arXiv:2306.15195}.

\bibitem{lin2014microsoftCOCO}
Lin, T.-Y., Maire, M., Belongie, S., Bourdev, L., Girshick, R., Hays, J., et~al. (2014).
\newblock Microsoft coco: Common objects in context.
\newblock {\em arXiv preprint arXiv:1405.0312}.

\bibitem{zhang2023exploringAD}
Zhang, J., Chen, X., Xue, Z., Wang, Y., Wang, C., Liu, Y. (2023).
\newblock Exploring grounding potential of vqa-oriented gpt-4v for zero-shot
  anomaly detection.
\newblock {\em arXiv preprint arXiv:2311.02612}.

\bibitem{tang2023generalizationcod}
Tang, L., Jiang, P.-T., Shen, Z., Zhang, H., Chen, J., Li, B. (2023).
\newblock Generalization and hallucination of large vision-language models
  through a camouflaged lens.
\newblock {\em arXiv preprint arXiv:2311.11273}.

\bibitem{10261199}
Qiu, J., Li, L., Sun, J., Peng, J., Shi, P., Zhang, R., et~al. (2023).
\newblock Large ai models in health informatics: Applications, challenges, and
  the future.
\newblock {\em IEEE Journal of Biomedical and Health Informatics},
  27(12):6074--6087.

\bibitem{gu2023anomalygpt}
Gu, Z., Zhu, B., Zhu, G., Chen, Y., Tang, M., et~al. (2023b).
\newblock Anomalygpt: Detecting industrial anomalies using large
  vision-language models.
\newblock {\em arXiv preprint arXiv:2308.15366}.

\bibitem{fu2023mme}
Fu, C., Chen, P., Shen, Y., Qin, Y., Zhang, M., Lin, X., et~al. (2023a).
\newblock Mme: A comprehensive evaluation benchmark for multimodal large
  language models. arxiv 2306.13394 (2023).
\newblock {\em arXiv preprint arXiv:2306.13394}.

\bibitem{song20233d}
Song, R., Zhang, W., Zhao, Y., Liu, Y., Rosin, P.~L. (2023).
\newblock 3d visual saliency: an independent perceptual measure or a derivative
  of 2d image saliency?
\newblock {\em IEEE Transactions on Pattern Analysis and Machine Intelligence},
  45(11):13083--13099.

\bibitem{fu2021siamese}
Fu, K., Fan, D.-P., Ji, G.-P., Zhao, Q., Shen, J., Zhu, C. (2021).
\newblock Siamese network for rgb-d salient object detection and beyond.
\newblock {\em IEEE Transactions on Pattern Analysis and Machine Intelligence},
  44(9):5541--5559.

\bibitem{fu2020jl}
Fu, K., Fan, D.-P., Ji, G.-P., Zhao, Q. (2020).
\newblock Jl-dcf: Joint learning and densely-cooperative fusion framework for
  rgb-d salient object detection.
\newblock In {\em Proceedings of the IEEE/CVF conference on computer vision and
  pattern recognition}, pages 3052--3062. Piscataway: IEEE.

\bibitem{xie2020segmentingTrans10K}
Xie, E., Wang, W., Wang, W., Ding, M., Shen, C., Luo, P. (2020).
\newblock Segmenting transparent objects in the wild.
\newblock In Vedaldi, A., Bischof, H., Brox, T., et~al. (Eds.), {\em
  Proceedings of the 16th European conference on computer vision}, pages
  696--711. Cham: Springer.

\bibitem{fan2022concealed}
Fan, D.-P., Ji, G.-P., Cheng, M.-M., Shao, L. (2021).
\newblock Concealed object detection.
\newblock {\em IEEE Transactions on Pattern Analysis and Machine Intelligence},
  44(10):6024--6042.

\bibitem{ji2022fast}
Ji, G.-P., Zhu, L., Zhuge, M., Fu, K. (2022).
\newblock Fast camouflaged object detection via edge-based reversible
  re-calibration network.
\newblock {\em Pattern Recognition}, 123:108414.

\bibitem{codella2018skinISIC2017}
Codella, N.~C., Gutman, D., Celebi, M.~E., Helba, B., Marchetti, M.~A., Dusza,
  S.~W., et~al. (2018).
\newblock Skin lesion analysis toward melanoma detection: A challenge at the
  2017 international symposium on biomedical imaging (isbi), hosted by the
  international skin imaging collaboration (isic).
\newblock In {\em Proceedings of the IEEE International Symposium on Biomedical
  Imaging}, pages 168--172. Piscataway: IEEE.

\bibitem{tajbakhsh2015automatedColonDB}
Tajbakhsh, N., Gurudu, S.~R., Liang, J. (2015).
\newblock Automated polyp detection in colonoscopy videos using shape and
  context information.
\newblock {\em IEEE Transactions on Medical Imaging}, 35(2):630--644.

\bibitem{bergmann2021mvtec}
Bergmann, P., Batzner, K., Fauser, M., Sattlegger, D., Steger, C. (2021).
\newblock The mvtec anomaly detection dataset: a comprehensive real-world
  dataset for unsupervised anomaly detection.
\newblock {\em International Journal of Computer Vision}, 129(4):1038--1059.

\bibitem{fan2018salientSOC}
Fan, D.-P., Cheng, M.-M., Liu, J.-J., Gao, S.-H., Hou, Q., Borji, A.
  (2018).
\newblock Salient objects in clutter: Bringing salient object detection to the
  foreground.
\newblock In Ferrari, V., M., H., C., S., et~al. (Eds.), {\em Proceedings
  of the 15th European conference on computer vision}, pages 186--202. Cham:
  Springer.

\bibitem{wang2020improvedchildb}
Wang, W., Tian, J., Zhang, C., Luo, Y., Wang, X., Li, J. (2020).
\newblock An improved deep learning approach and its applications on colonic
  polyp images detection.
\newblock {\em BMC Medical Imaging}, 20:1--14.

\bibitem{zou2022spotVisA}
Zou, Y., Jeong, J., Pemula, L., Zhang, D., Dabeer, O. (2022).
\newblock Spot-the-difference self-supervised pre-training for anomaly
  detection and segmentation.
\newblock In Avidan, S., Brostow, G., Ciss{\'{e}}, M., et~al. (Eds.), {\em
  Proceedings of the 17th European conference on computer vision}, pages
  392--408. Cham: Springer.

\bibitem{wang2017learningDUTS}
Wang, L., Lu, H., Wang, Y., Feng, M., Wang, D., Yin, B., et~al. (2017).
\newblock Learning to detect salient objects with image-level supervision.
\newblock In {\em Proceedings of the IEEE conference on computer vision and
  pattern recognition}, pages 136--145. Piscataway: IEEE.

\bibitem{silva2014towardETIS}
Silva, J., Histace, A., Romain, O., Dray, X., Granado, B. (2014).
\newblock Toward embedded detection of polyps in wce images for early diagnosis
  of colorectal cancer.
\newblock {\em International Journal of Computer Assisted Radiology and
  Surgery}, 9:283--293.

\bibitem{conti2023vocabularyfree}
Conti, A., Fini, E., Mancini, M., Rota, P., Wang, Y., Ricci, E. (2023).
\newblock Vocabulary-free image classification.
\newblock In OH, A., Naumann, T., Globerson, A., et~al. (Eds.), {\em
  Proceedings of the 37th international conference on neural information
  processing systems}, pages 30662--30680. Red Hook: Curran Associates.

\bibitem{li2023evaluatingobjecthallucination}
Li, Y., Du, Y., Zhou, K., Wang, J., Zhao, W.~X., Wen, J.-R. (2023b).
\newblock Evaluating object hallucination in large vision-language models.
\newblock {\em arXiv preprint arXiv:2305.10355}.

\bibitem{xu2023lvlm}
Xu, P., Shao, W., Zhang, K., Gao, P., Liu, S., Lei, M., et~al. (2023).
\newblock Lvlm-ehub: A comprehensive evaluation benchmark for large
  vision-language models.
\newblock {\em arXiv preprint arXiv:2306.09265}.

\bibitem{cui2023holistic}
Cui, C., Zhou, Y., Yang, X., Wu, S., Zhang, L., Zou, J., et~al. (2023).
\newblock Holistic analysis of hallucination in gpt-4v (ision): Bias and
  interference challenges.
\newblock {\em arXiv preprint arXiv:2311.03287}.

\bibitem{kirillov2023segmentanything}
Kirillov, A., Mintun, E., Ravi, N., Mao, H., Rolland, C., Gustafson, L., et~al. (2023).
\newblock Segment anything.
\newblock In {\em Proceedings of the IEEE/CVF international conference on
  computer vision}, pages 4015--4026. Piscataway: IEEE.

\bibitem{dosovitskiy2020imageVIT}
Dosovitskiy, A., Beyer, L., Kolesnikov, A., Weissenborn, D., Zhai, X., Unterthiner, T., et~al. (2021).
\newblock An image is worth 16x16 words: Transformers for image recognition at scale.
\newblock In {\em Proceedings of the 9th International Conference on Learning Representations}, pages 1-21, Retrieved June 4, 2024, from https://openreview.net/forum?id=YicbFdNTTy, 2021.

\bibitem{electronics10030279}
Padilla, R., Passos, W.~L., Dias, T.~L., Netto, S.~L., Da~Silva, E.~A. (2021).
\newblock A comparative analysis of object detection metrics with a companion
  open-source toolkit.
\newblock {\em Electronics}, 10.

\bibitem{Perazzi2012SaliencyFC}
Perazzi, F., Kr{\"a}henb{\"u}hl, P., Pritch, Y., Hornung, A. (2012).
\newblock Saliency filters: Contrast based filtering for salient region
  detection.
\newblock In {\em Proceedings of the IEEE conference on computer vision and
  pattern recognition}, pages 733--740. Piscataway: IEEE.

\bibitem{Fan2017StructureMeasureAN}
Fan, D.-P., Cheng, M.-M., Liu, Y., Li, T., Borji, A. (2017).
\newblock Structure-measure: A new way to evaluate foreground maps.
\newblock In {\em Proceedings of the IEEE international conference on computer
  vision}, pages 4558--4567. Piscataway: IEEE.

\bibitem{Achanta2009FrequencytunedSR}
Achanta, R., Hemami, S., Estrada, F., Susstrunk, S. (2009).
\newblock Frequency-tuned salient region detection.
\newblock In {\em Proceedings of the IEEE conference on computer vision and
  pattern recognition}, pages 1597--1604. Piscataway: IEEE.

\bibitem{gu2023systematicpromptengineering}
Gu, J., Han, Z., Chen, S., Beirami, A., He, B., Zhang, G., et~al. (2023a).
\newblock A systematic survey of prompt engineering on vision-language
  foundation models.
\newblock {\em arXiv preprint arXiv:2307.12980}.

\bibitem{li2024llavamed}
Li, C., Wong, C., Zhang, S., Usuyama, N., Liu, H., Yang, J., et~al. (2023a).
\newblock Llava-med: Training a large language-and-vision assistant for
  biomedicine in one day.
\newblock In OH, A., Naumann, T., Globerson, A., et~al. (Eds.), {\em
  Proceedings of the 37th international conference on neural information
  processing systems}, pages 28541--28564. Red Hook: Curran Associates.

\bibitem{liu2023promoting}
Liu, X., Fu, K., Zhao, Q. (2023c).
\newblock Promoting segment anything model towards highly accurate dichotomous
  image segmentation.
\newblock {\em arXiv preprint arXiv:2401.00248}.

\bibitem{zhou2023analyzingobjecthallucination}
Zhou, Y., Cui, C., Yoon, J., Zhang, L., Deng, Z., Finn, C., et~al. (2023).
\newblock Analyzing and mitigating object hallucination in large
  vision-language models.
\newblock {\em arXiv preprint arXiv:2310.00754}.

\bibitem{qian2024easyhallucination}
Qian, Y., Zhang, H., Yang, Y., Gan, Z. (2024).
\newblock How easy is it to fool your multimodal llms? an empirical analysis on
  deceptive prompts.
\newblock {\em arXiv preprint arXiv:2402.13220}.

\bibitem{kim2023exposinghallucination}
Kim, J.~M., Koepke, A., Schmid, C., Akata, Z. (2023).
\newblock Exposing and mitigating spurious correlations for cross-modal
  retrieval.
\newblock In {\em Proceedings of the IEEE/CVF conference on computer vision and
  pattern recognition}, pages 2584--2594. Piscataway: IEEE.

\bibitem{wu2023improvingCrossTaskGeneralization}
Wu, Y., Zhao, Y., Li, Z., Qin, B., Xiong, K. (2023).
\newblock Improving cross-task generalization with step-by-step instructions.
\newblock {\em Science China Information Sciences}. Advance online publication. \url{https://doi.org/10.1007/s11432-023-3911-2}

\bibitem{chen2023feedback}
Chen, H., Yuan, K., Huang, Y., Guo, L., Wang, Y., Chen, J. (2023a).
\newblock Feedback is all you need: from chatgpt to autonomous driving.
\newblock {\em Science China Information Sciences}, 66(6):1--3.

\bibitem{yan2024vigor}
Yan, S., Bai, M., Chen, W., Zhou, X., Huang, Q., Li, L.~E. (2024).
\newblock Vigor: Improving visual grounding of large vision language models
  with fine-grained reward modeling.
\newblock {\em arXiv preprint arXiv:2402.06118}.

\bibitem{jiao2024enhancing}
Jiao, Q., Chen, D., Huang, Y., Li, Y., Shen, Y. (2024).
\newblock Enhancing multimodal large language models with vision detection
  models: An empirical study.
\newblock {\em arXiv preprint arXiv:2401.17981}.

\bibitem{yao2023deepspeed}
Yao, Z., Wu, X., Li, C., Zhang, M., Qi, H., Ruwase, O., et~al. (2023).
\newblock Deepspeed-visualchat: Multi-round multi-image interleave chat via
  multi-modal causal attention.
\newblock {\em arXiv preprint arXiv:2309.14327}.

\bibitem{fu2022light}
Fu, K., Jiang, Y., Ji, G.-P., Zhou, T., Zhao, Q., Fan, D.-P. (2022).
\newblock Light field salient object detection: A review and benchmark.
\newblock {\em Computational Visual Media}, 8(4):509--534.

\bibitem{he2022fewshot}
He, J., Fu, K. (2022).
\newblock Rgb-d salient object detection of using few-shot learning.
\newblock {\em Journal of Image and Graphics}, 27(10):2860--2872.

\bibitem{zhou2021specificity}
Zhou, T., Fu, H., Chen, G., Zhou, Y., Fan, D.-P., Shao, L. (2021).
\newblock Specificity-preserving rgb-d saliency detection.
\newblock In {\em Proceedings of the IEEE/CVF international conference on
  computer vision}, pages 4681--4691. Piscataway: IEEE.

\bibitem{chen2021rgb}
Chen, Q., Liu, Z., Zhang, Y., Fu, K., Zhao, Q., Du, H. (2021).
\newblock Rgb-d salient object detection via 3d convolutional neural networks.
\newblock In {\em Proceedings of the 35nd AAAI conference on artificial
  intelligence}, pages 1063--1071. Palo Alto: AAAI Press.

\bibitem{fu2019deepside}
Fu, K., Zhao, Q., Gu, I. Y.-H., Yang, J. (2019).
\newblock Deepside: A general deep framework for salient object detection.
\newblock {\em Neurocomputing}, 356:69--82.

\bibitem{zhang2021depth}
Zhang, W., Ji, G.-P., Wang, Z., Fu, K., Zhao, Q. (2021).
\newblock Depth quality-inspired feature manipulation for efficient rgb-d
  salient object detection.
\newblock In Shen, H.~T., Zhuang, Y., Smith, J., et~al. (Eds.), {\em
  Proceedings of the 29th ACM International Conference on Multimedia}, pages
  731--740. New York: ACM.

\bibitem{zhong2024vlm}
Zhong, L., Liao, X., Zhang, S., Zhang, X., Wang, G. (2024).
\newblock Vlm-cpl: Consensus pseudo labels from vision-language models for
  human annotation-free pathological image classification.
\newblock {\em arXiv preprint arXiv:2403.15836}.

\bibitem{wang2022language}
Wang, Z., Li, M., Xu, R., Zhou, L., Lei, J., Lin, X., et~al. (2022).
\newblock Language models with image descriptors are strong few-shot
  video-language learners.
\newblock {\em Proceedings of the 36th international conference on neural
  information processing systems}, 35:8483--8497.

\bibitem{he2024decouplingDshmp}
He, S., Ding, H. (2024).
\newblock Decoupling static and hierarchical motion perception for referring
  video segmentation.
\newblock {\em arXiv preprint arXiv:2404.03645}.

\bibitem{ding2023mevisMeVis}
Ding, H., Liu, C., He, S., Jiang, X., Loy, C.~C. (2023a).
\newblock Mevis: A large-scale benchmark for video segmentation with motion
  expressions.
\newblock In {\em Proceedings of the IEEE/CVF international conference on
  computer vision}, pages 2694--2703. Piscataway: IEEE.

\bibitem{ding2023moseMoSE}
Ding, H., Liu, C., He, S., Jiang, X., Torr, P.~H., Bai, S. (2023b).
\newblock Mose: A new dataset for video object segmentation in complex scenes.
\newblock In {\em Proceedings of the IEEE/CVF international conference on
  computer vision}, pages 20224--20234. Piscataway: IEEE.

\bibitem{zhang2019reconstruct}
Zhang, W., Wang, B., Ma, L., Liu, W. (2019).
\newblock Reconstruct and represent video contents for captioning via
  reinforcement learning.
\newblock {\em IEEE Transactions on Pattern Analysis and Machine Intelligence},
  42(12):3088--3101.

\end{thebibliography}
% }

% \vfill

\end{document}